\pdfoutput=1
\documentclass[11pt]{article}

\usepackage[preprint]{acl}

\usepackage{times}
\usepackage{latexsym}
\usepackage{amsmath}

\usepackage[T1]{fontenc}
\usepackage[utf8]{inputenc}

\usepackage{microtype}

\usepackage{inconsolata}

\usepackage{graphicx}
\usepackage{svg}

\usepackage{booktabs} % For formal tables

\title{Emergence of Episodic Memory in Transformers: Characterizing Changes in Temporal Structure of Attention Scores During Training}

\author{Deven Mahesh Mistry, \textbf{Anooshka Bajaj}, \textbf{Yash Aggarwal}, \\ \textbf{Sahaj Singh Maini}, \textbf{Zoran Tiganj}\\
        Department of Computer Science\\ Indiana University Bloomington}
\begin{document}
\maketitle
\begin{abstract}
We investigate in-context temporal biases in attention heads and transformer outputs. Using cognitive science methodologies, we analyze attention scores and outputs of the GPT-2 models of varying sizes. Across attention heads, we observe effects characteristic of human episodic memory,  including temporal contiguity, primacy and recency. Transformer outputs demonstrate a tendency toward in-context serial recall. Importantly, this effect is eliminated after the ablation of the induction heads, which are the driving force behind the contiguity effect.
Our findings offer insights into how transformers organize information temporally during in-context learning, shedding light on their similarities and differences with human memory and learning.
\end{abstract}

\section{Introduction}
Large language models (LLMs) have demonstrated a remarkable capacity for in-context learning. They are capable of adapting to new tasks using examples provided within the input prompt, without any parameter updates \cite{brown2020language}. The temporal position of tokens plays an important role in in-context learning. For instance, simply asking the model to repeat a sequence of words from the prompt requires the model to use the information about the temporal position of tokens. This resembles human learning, where the temporal organization of memory plays a critical role in recalling specific past episodes.

Previous work has demonstrated that some attention heads show a temporal induction property. These \textit{induction heads} search the input prompt for the prior occurrence of the current token. If a match is found, they attend to the token that followed the previous presentation of the current token. This mechanism allows induction heads to effectively learn and reproduce sequences of tokens and it has been argued to contribute to the model's ability to perform tasks based on contextual information \cite{olsson2022context,elhage2021mathematical,singh2024needs, ji2024linking, pink2024assessingepisodicmemoryllms}. 

The induction property is related to human episodic memory. Numerous studies have demonstrated that episodic memory exhibits contiguity effect, where items or events that occur close together in time are more likely to be remembered together in memory recall \cite{kahana1996associative,howard2002contextual,lohnas2021role,polyn2009context,jenkins2010prefrontal}. For example, if a person experiences a sequence of events within a short time span, they are more likely to recall these events together than if they were spread out over a longer period. This effect was systematically studied through free recall, a task where participants are presented with a list of items (e.g., words, pictures) and then asked to recall them in any order they choose. Participants tend to recall items that were presented close together in the original list in clusters, indicating that the temporal proximity during encoding influences the retrieval process. This supports the idea that our memories are organized not just by the content of a stimulus but also by its temporal context \cite{howard2002contextual,polyn2009context}. Neuroscience studies have demonstrated that neural activity in the brain during recall reinstates the neural activity observed during encoding, indicating the retrieval of temporal context or a mental ``jump back in time'' \cite{howard2012ensembles,folkerts2018human}. The contiguity effect has been recently studied in deep neural networks, including recurrent neural networks \cite{limodeling} and pretrained transformer models \cite{ji2024linking} where it was found that pretrained transformers contain attention heads that recover temporal context in a manner consistent with human episodic memory. Furthermore, human-like episodic memory is found to improve performance in tasks that require processing over extended temporal context \cite{fountas2024human}.

In addition to temporal induction, transformers exhibit \textit{serial position effects}, specifically \textit{recency} and \textit{primacy} \cite{janik2023aspects,peysakhovich2023attention,wang2023primacy,guo2024serial,angne2023two}. Recency implies that tokens that are more recent in the prompt (closer to the present token) are going to be more important in generating the subsequent token. Conversely, primacy implies the same property but for the tokens that are presented at the beginning of the prompt. The emergence of these effects in transformers can be shaped by positional encoding \cite{janik2023aspects,peysakhovich2023attention}. Positional encoding embeddings can be learned \cite{devlin2018bert} or fixed to vectors that convey positional information, such as Rotformer, which uses rotary positional embeddings \cite{su2024roformer} or early transformer models that used sinusoidal embeddings \cite{vaswani2017attention}. These positional embeddings capture translation invariance, monotonicity and symmetry through distance metrics in their vector embeddings \cite{wang2021on}.

Recency and primacy are also characteristic of human memory, where we often exhibit better recall for items presented at the beginning (primacy effect) and end (recency effect) of a list \cite{Atkinson1968,Glanzer1966,Murdock1962}. This parallel suggests that LLMs, like humans, may be sensitive to the temporal context of information, with recent and initial tokens playing a more significant role in shaping the model's internal representations and influencing its output. 

Here we investigate the temporal aspects of attention heads and transfromer outputs during learning. We train two transformer models of different sizes (GPT-2 small and GPT-2 medium) on three datasets: Wikitext-103  \cite{Merity2016} and two sampled datasets of FineWeb \cite{huggingfacefw_2024} (with 1B tokens and 10B tokens). We adopt tools used in cognitive science to characterize temporal aspects of human memory and use them to examine attention heads and outputs of transformer models. Specifically, we use Lag-Conditional Recall Probability  (Lag-CRP) analysis which measures the probability of recalling an item a certain number of positions away (called \textit{lag}) from the previously recalled item. We apply this analysis to transformer outputs and attention heads to understand how temporal relationships shape attention scores and token predictions.
We characterize the impact of trainable positional encoding, model size, and number of training interactions on the emergence of contiguity effect and serial position effects. 
We also ablate induction heads to examine their role in shaping temporal properties of transformer outputs. Our findings provide novel insights into similarities and differences between human memory and transformers, improving our understanding of in-context learning.

\section{Methods}
\subsection{Models and training}

We used models based on the GPT-2 small and GPT-2 medium architectures \cite{Radford2019LanguageMA}. GPT-2 small has approximately 124 million parameters, consisting of 12 attention heads, 12 transformer layers, an embedding size of 768 dimensions, and an MLP with 3,072 neurons. GPT-2 medium has approximately 353 million parameters and consists of 16 attention heads, 24 transformer layers, an embedding size of 1024 dimensions and an MLP with 4096 neurons. The vocabulary consists of 50,257 tokens and we used the GPT-2 byte-pair encoder.

We trained GPT-2 small and medium on Wikitext-103 \cite{Merity2016} for 4000 iterations. The dataset contains 117M tokens in the training set, 0.24M tokens in the validation set and 0.28M tokens in the test set. We also trained GPT-2 small on two larger datasets sampled from FineWeb \cite{huggingfacefw_2024} that included 1B tokens and 10B tokens. 1B dataset had 996M tokens in the training set and 4M in the validation set and 10B dataset had 10.3B tokens in the training set and 51.4M in the validation set. 
In our experiments, we used the \textit{nanoGPT} codebase \cite{nanoGPT}. We trained all the configurations of GPT-2 small and GPT-2 medium with the same hyperparameters, including learning rate, number of warm-up iterations, and weight decay. During training, all models had a maximum learning rate of $10^{-4}$ and a learning rate warm-up period of 450 iterations. The training was done on four 40GB A100 GPUs.

\subsection{Calculation of lag-CRP curves for attention heads}
To compute lag-CRP curves for attention heads, we prompt the models with a 1000 tokens long prompt composed of source and destination sequences (the maximum length of the prompt for both GPT-2 small and GPT-2 medium is 1024 tokens). The source sequence consisted of 500 most frequent tokens in a given dataset. The tokens were presented in a random order. The source sequence was followed by 500 tokens long destination sequence. The tokens in the destination sequence were the exact repetition of the source sequence. We computed the attention scores between each token in the destination sequence and the source sequence to calculate the lag-CRP curve. Given a single sequence of tokens, the lag-CRP score for lag at position zero was calculated as the average attention score between the same tokens in the source and destination sequences. Similarly, the lag-CRP score for lag $l$ was calculated as the average attention score between the tokens in the destination sequence, and a different token placed $l$ positions away from the corresponding token in the source sequence. When $l$ is positive, the lag-CRP score is calculated for the tokens following the corresponding token in the source sequence, and when $l$ is negative, the score is calculated for the tokens that occur before the corresponding token. To reduce the impact of semantic similarity, we averaged the lag-CRP scores for a given lag across ten randomly permuted sequences.  We produce the lag-CRP curves for all heads. Mathematically, the lag-CRP score for an individual head is computed as follows \cite{ji2024linking}:
% \tilde{\alpha}_{\text{l}}
\begin{equation}
   S_{l} = \sum_{i\in M}{\frac{1}{N - |l| * 2} \sum_{|l|<s \leq N-|l|} a_{s+N,s+l}}.
\end{equation}

Here $S_l$ refers to the score for lag $l$, $N$ is the size of the source (or) destination sequence, $M$ is the set of example sequences generated by permuting the source tokens along with similarly permuted destination tokens. $a_{i, j}$ is the attention score calculated between the token at position $i$ in the destination sequence and position $j$ in source sequence. In all our experiments, $M$ contains 10 sequences. We note that since we used attention scores to compute the lag-CRP curve, it is no longer restricted to response probabilities, therefore it can take any range of values.

\subsection{Calculation of induction matching score}
Following previous work on in-context learning \cite{olsson2022context,elhage2021mathematical} we computed induction scores for each attention head. The induction score expresses the degree to which the head attends to the token following the previous occurrence of the current token in the sequence. 

Given a sequence of tokens to compute the induction matching score for an attention head, we first extract the attention pattern from the model for the corresponding layer and head. This attention pattern provides the weights indicating how much each token in the sequence attends to every other token. We then construct a target matrix that records matches based on the induction rule: if the token at a destination position matches the token before a particular source position, the corresponding entry in the target matrix is set to 1. Next, we compute the element-wise product of the attention pattern and the target matrix to isolate attention values corresponding to induction matches. The numerator is the sum of these matched attention values, while the denominator is the sum of all attention values between the source and destination positions. The induction matching score is obtained by dividing the numerator by the denominator.

More formally, let $N$ be the sequence length, $a_{i,j}$ denote the attention value from token $i$ (destination position) to token $j$ (source position), and $t_{i,j}$ be the target matrix entry, where $t_{i,j} = 1$ if the token at position $i$ matches the token before position $j$, and $t_{i,j} = 0$ otherwise. The induction matching score is given by:
\begin{equation}
 \text{I} = \frac{\sum_{(i,j)} a_{i,j} \cdot t_{i,j}}{\sum_{(i,j)} a_{i,j}},
\end{equation}
where the summation $\sum_{(i,j)}$ is taken over all valid pairs of $(i,j)$ within the sequence.

\subsection{Computing the temporal extent of the contiguity effect and strength of recency effect}
While the induction matching score quantifies the tendency to attend to the token that follows the previous occurrence of the current token, human episodic memory is characterized by a lag-CRP curve that has a strong contiguity effect. This implies a gradual falloff of the lag-CRP curve as a function of positive and negative lags. In our experiments, we choose a subset of heads that have the highest lag-CRP score at $l=1$ when the lag-CRP curve is computed between $l=-10$ to $l=10$. This enabled the identification of temporal contiguity even in the presence of high recency or primacy (i.e., high values of the lag-CRP curve for large values of $l$).

To quantify the recency effect, we computed the lag-CRP curves for the chosen heads. To isolate the recency effect and to ensure that the contiguity effect does not superimpose with the recency effect, we remove the lag-CRP scores between the lags -50 and +50. We then fit a linear regression model to the rest of the lag-CRP scores and compute the average slope of the linear fits across selected attention heads. This expresses the recency bias in the attention heads. 

In our experiments, while exploring the contiguity effect in the selected attention heads, we observed a gradual falloff of lag-CRP scores as a function of positive but not negative lags. To quantify this falloff, we first subtract the linear fit from the lag-CRP scores in order to remove the recency effect. We then fit an exponential function to the positive lags of the lag-CRP curve (exponential fit is commonly applied in human episodic memory models \cite{howard2002contextual,polyn2009context}). We used Levenberg-Marquardt algorithm \cite{marquardt1963algorithm, levenberg1944method} and optimized the following function: $ae^{-t/\tau}$ where $a$ and $\tau$ (time constant) are parameters and $t$ is the lag (Tab. \ref{tab:training_iteration_wikitext103}).

\subsection{Positional encoding with different magnitudes}
We trained five variants of the GPT-2 small model with five different magnitudes of positional encoding:  0, 0.5, 1, 1.5, and 2. These factors multiplied positional embeddings before positional embeddings were added to word embeddings. Positional embeddings were learnable, and they used the same weight initialization.

\section{Results}
We trained GPT-2 small and GPT-2 medium transformer models. On the Wikitext-103 dataset GPT-2 small converged after around 4000 iterations, and the larger GPT-2 medium model converged after around 2000 iterations. The convergence was determined by monitoring the validation loss. The models converged to perplexities similar to those of models with comparable size, including Transformer-XL Standard \cite{dai-etal-2019-transformer} LaMemo \cite{ji-etal-2022-lamemo}, Hybrid H3  \cite{fu2023hungry} and TrimeLM Long \cite{zhong-etal-2022-training}, all of which have perplexity above 20 (Tab.~\ref{tab:perplexity}). Importantly, regardless of the magnitude of the positional encoding, the models converged to similar values of perplexity, consistent with results in \citet{haviv2022transformer}. We also trained GPT-2 small on FineWeb-1B and 10B. Both models converged after around 10000 iterations. 

\begin{table*}[h!]
\centering
\caption{Lowest perplexity values for GPT-2 small and  GPT-2 medium for different positional encoding magnitudes after training on Wikitext-103 dataset.}
\label{tab:perplexity}
\begin{tabular}{@{}lccccccc@{}} % 'l' for the metric name column and 'c' for the center-aligned data columns
\toprule
 & \multicolumn{5}{c}{GPT-2 small} & \multicolumn{1}{c}{GPT-2 medium} \\
\cmidrule(lr){2-6} \cmidrule(lr){7-7} % Corrected cmidrule to accurately reflect the header spans
Positional encoding magnitude & 0 & 0.5 & 1 & 1.5 & 2  & 1 \\ \midrule
Perplexity & 19.8 & 19.8 & 19.8 & 19.7 & 19.8  & 19.5 \\ \bottomrule
\end{tabular}
\end{table*}

\begin{figure*}[h!]
    \centering
    \includegraphics[width=1\textwidth]{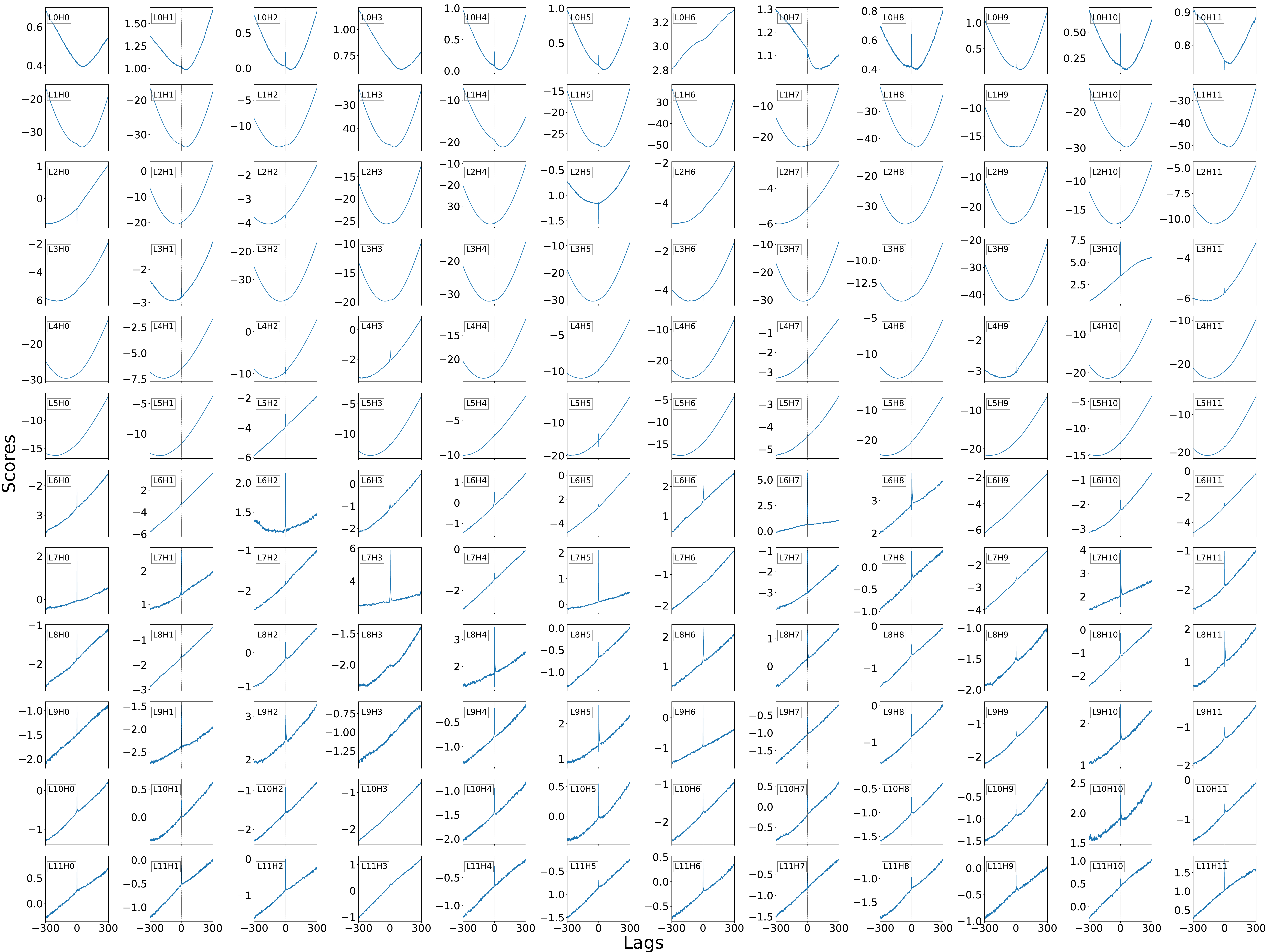}
    \caption{Attention scores as a function of lag for all attention heads in GPT-2 small after 4000 iterations on the Wikitext-103 dataset (baseline positional encoding). \label{fig:crp_all_GPT2-small_4000_1}} 
\end{figure*}

\begin{figure}[h!]
    \centering
    \begin{tabular}{ll}
    \textbf{A} &
    \textbf{B} \\
    {\includegraphics[width=0.44\columnwidth]{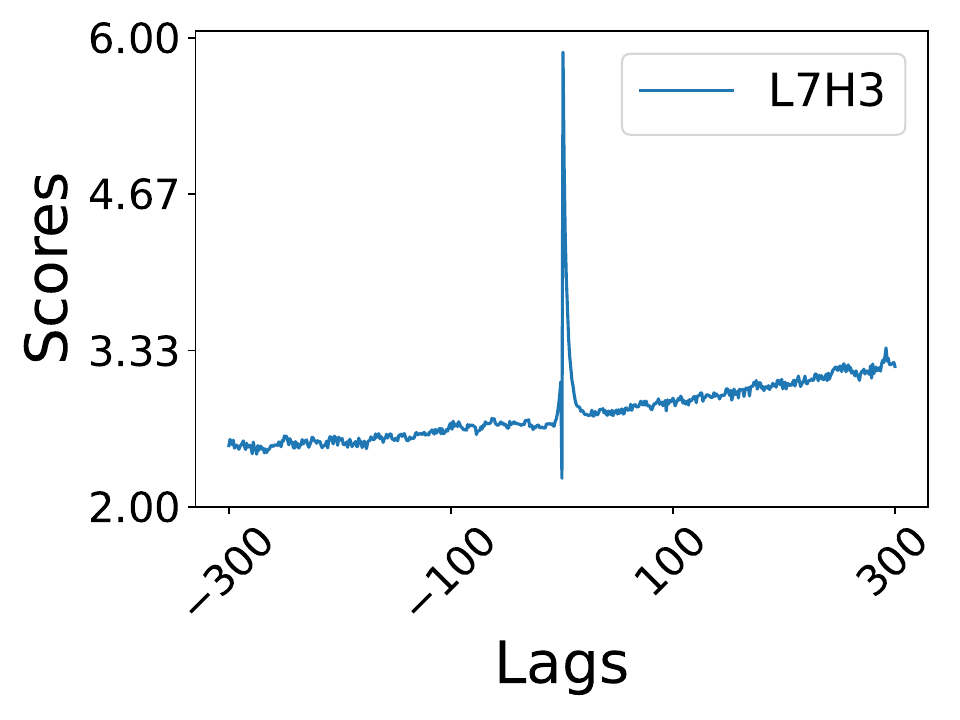}} &
    {\includegraphics[width=0.440\columnwidth]{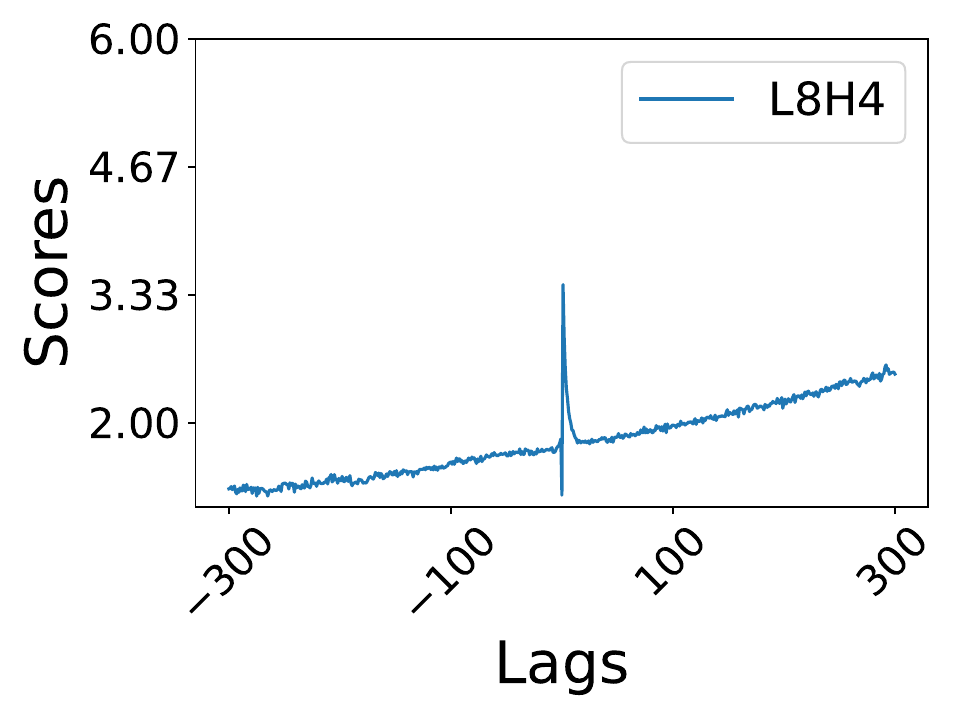}}\\
    \end{tabular}
    \begin{tabular}{ll}
    \textbf{C} &
    \textbf{D}  \\
    {\includegraphics[width=0.44\columnwidth]{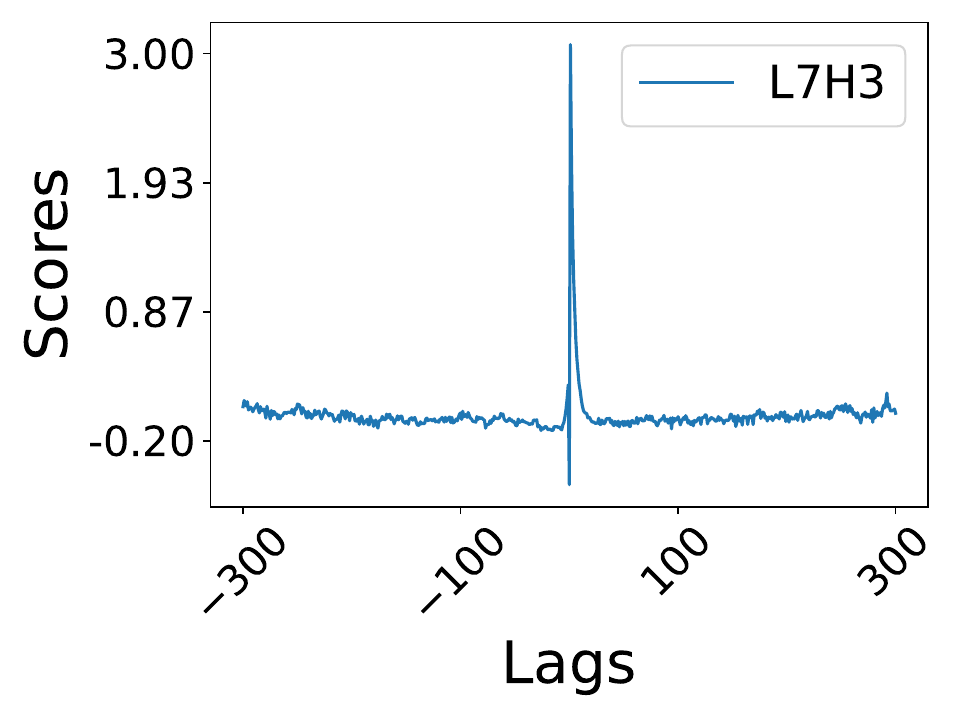}}  &{\includegraphics[width=0.44\columnwidth]{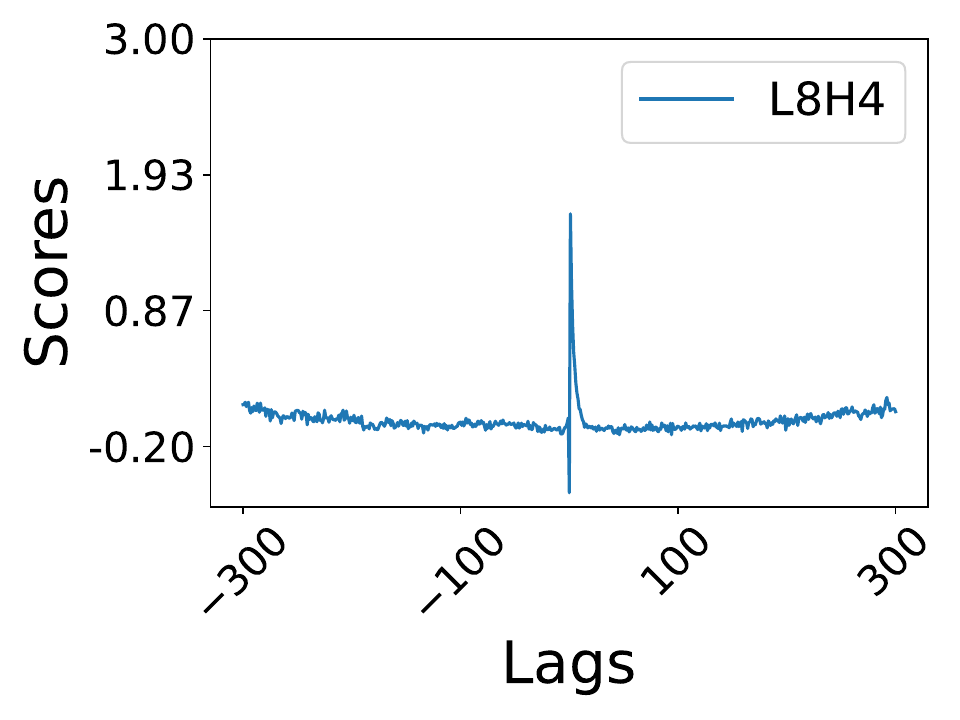}}\\
    \end{tabular}
    \begin{tabular}{ll}
    \textbf{E} &
    \textbf{F}  \\
    {\includegraphics[width=0.44\columnwidth]{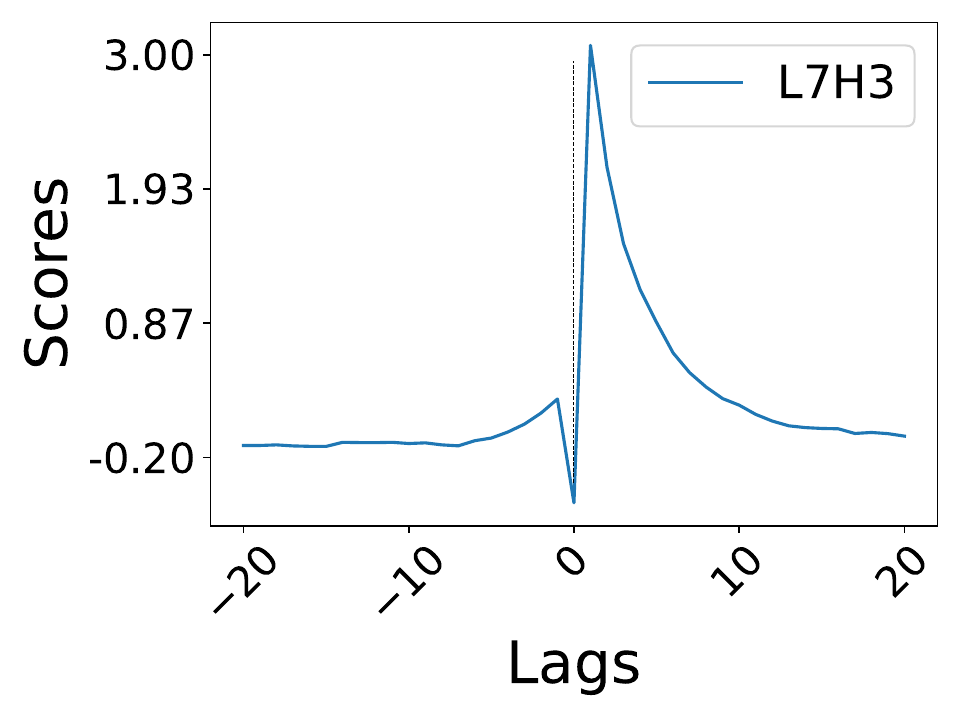}} &
    {\includegraphics[width=0.44\columnwidth]{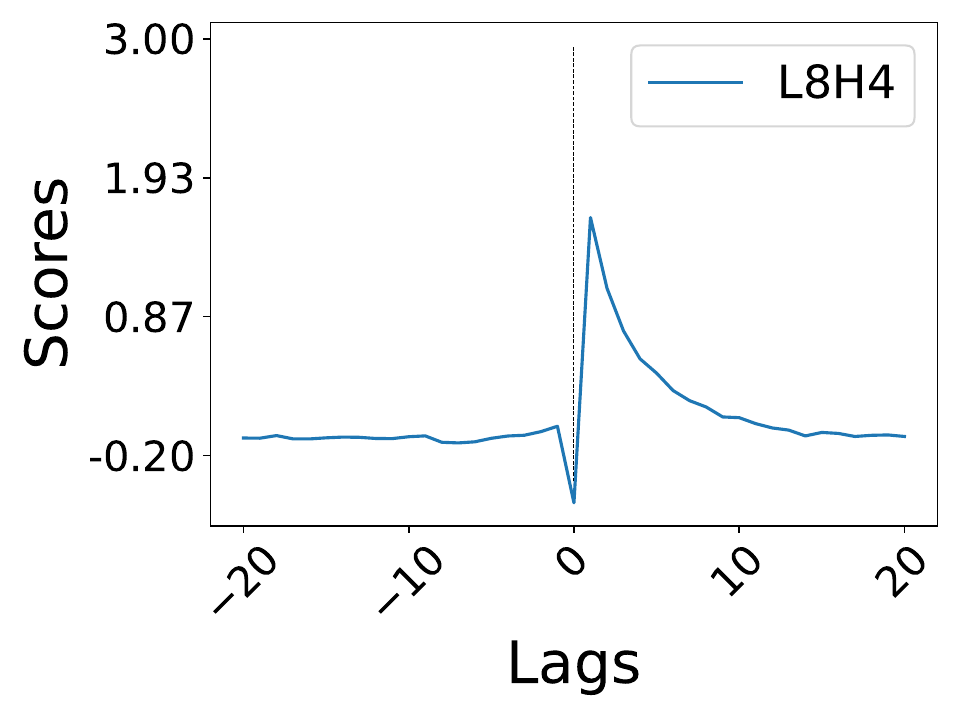}}\\
    \end{tabular}
    \caption{Two induction heads before (top row) and after (middle row) adjusting for recency effect. The bottom row shows zoomed-in version of the middle row.\label{fig:crp}}
\end{figure}
After training, most attention heads exhibited some form of structured temporal modulation--recency, primacy, contiguity, or a combination of these effects. For Wikitext-103, heads in layers closer to the input were characterized by recency and primacy effects, while heads in layers closer to the output had strong recency and contiguity effects. Attention scores of GPT-2 small as a function of lag across all heads before and after training are shown in Fig.~\ref{fig:crp_all_GPT2-small_0_1}  and Fig.~\ref{fig:crp_all_GPT2-small_4000_1} respectively. Attention scores across all heads after training of GPT-2 medium model are shown in Fig.~\ref{fig:crp_all_GPT-medium_2000_1}. Fig.~\ref{fig:crp} shows attention scores as a function of lags for two representative attention heads from layers closer to the output. The top row (Fig.~\ref{fig:crp}A-B) illustrates the raw values, while the middle and bottom rows (Fig.~\ref{fig:crp}C-F) show results after the recency effect has been removed, highlighting the contiguity effect. For FineWeb-1B and 10B attention scores across all heads after training are shown in Fig.~\ref{fig:crp_all_GPT2-small_1000_1B} and Fig.~\ref{fig:crp_all_GPT2-small_1000_10B} respectively. While similar to the results on Wikitext-103, they often showed more complex, non-linear patterns across a wide range of lags.

\subsection{Temporal properties of attention heads emerge gradually throughout training}

\begin{table*}[h!]
\centering
\caption{Induction head properties as a function of training iteration for the Wikitext-103 dataset.}
\label{tab:training_iteration_wikitext103}
\begin{tabular}{@{}lccccccc@{}} % 'l' for metric name and 'c' for data columns
\toprule
\multicolumn{1}{c}{} & \multicolumn{7}{c}{Training Iteration} \\
\cmidrule(l){2-8} 
Metric & 0 & 100 & 500 & 1000 & 2000 & 3000 & 4000 \\ 
\midrule
Average Induction Score & 0.0007 & 0.0007 & 0.0006 & 0.0012 & 0.0019 & 0.0015 & 0.0024 \\
Average Time Constant & 0.7 & 0.5 & 6.43 & 5 & 6.4 & 3.2 & 3.1 \\
Average Recency Slope & 0 & 0 & 0.0001 & 0.0015 & 0.0018 & 0.0032 & 0.0029 \\
Number of Induction Heads & 4 & 4 & 3 & 19 & 16 & 19 & 20 \\
\bottomrule
\end{tabular}
\end{table*}

\begin{table*}[h!]
\centering
\caption{Induction head properties as a function of training iteration for FineWeb-1B dataset.}
\label{tab:training_iteration_1B}
\begin{tabular}{@{}lccccccc@{}} % 'l' for metric name and 'c' for data columns
\toprule
\multicolumn{1}{c}{} & \multicolumn{7}{c}{Training Iteration} \\
\cmidrule(l){2-8} 
Metric & 0 & 50 & 100 & 500 & 1000 & 5000 & 10000 \\ 
\midrule
Average Induction Score & 0.0007 & 0.0007 & 0.0007 & 0.0002 & 0.008 & 0.011 & 0.016 \\
Number of Induction Heads & 4 & 4 & 1 & 3 & 25 & 36 & 25 \\
\bottomrule
\end{tabular}
\end{table*}

\begin{table*}[h!]
\centering
\caption{Induction head properties as a function of training iteration for FineWeb-10B dataset.}
\label{tab:training_iteration_10B}
\begin{tabular}{@{}lccccccc@{}} % 'l' for metric name and 'c' for data columns
\toprule
\multicolumn{1}{c}{} & \multicolumn{7}{c}{Training Iteration} \\
\cmidrule(l){2-8} 
Metric & 0 & 50 & 100 & 500 & 1000 & 5000 & 10000 \\ 
\midrule
Average Induction Score & 0.0007 & 0.0007 & 0.0007 & 0.0003 & 0.03 & 0.05 & 0.05 \\
Number of Induction Heads & 4 & 7 & 3 & 4 & 24 & 29 & 39 \\
\bottomrule
\end{tabular}
\end{table*}

\begin{figure}[h!]
    \centering
    \begin{tabular}{ll}
    \textbf{A} &
    \textbf{B}  \\
    {\includegraphics[width=0.44\columnwidth]{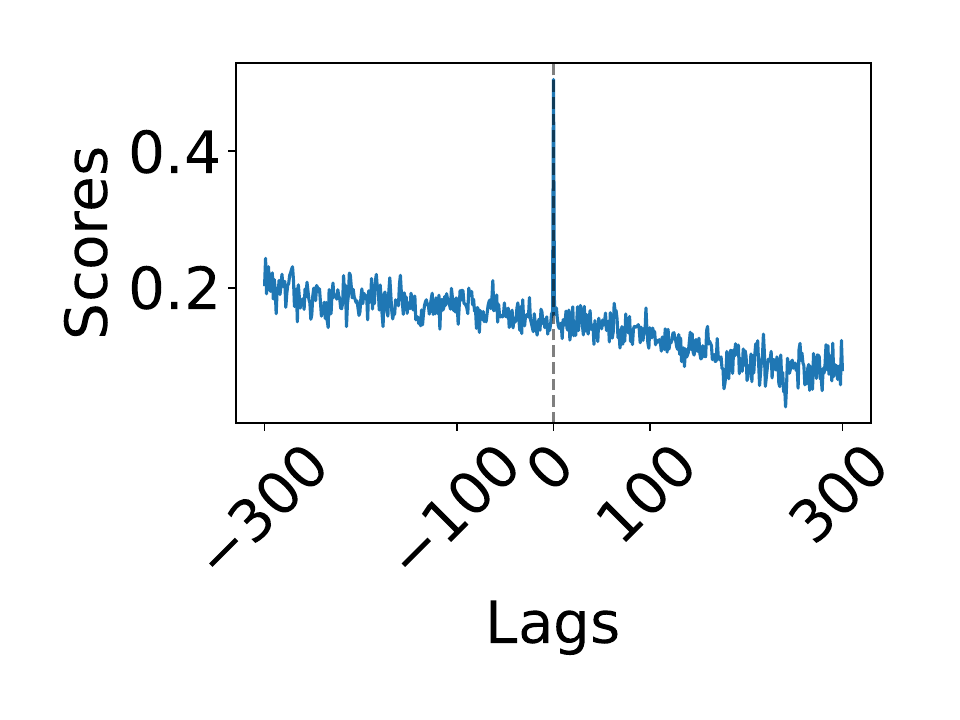}} &{\includegraphics[width=0.44\columnwidth]{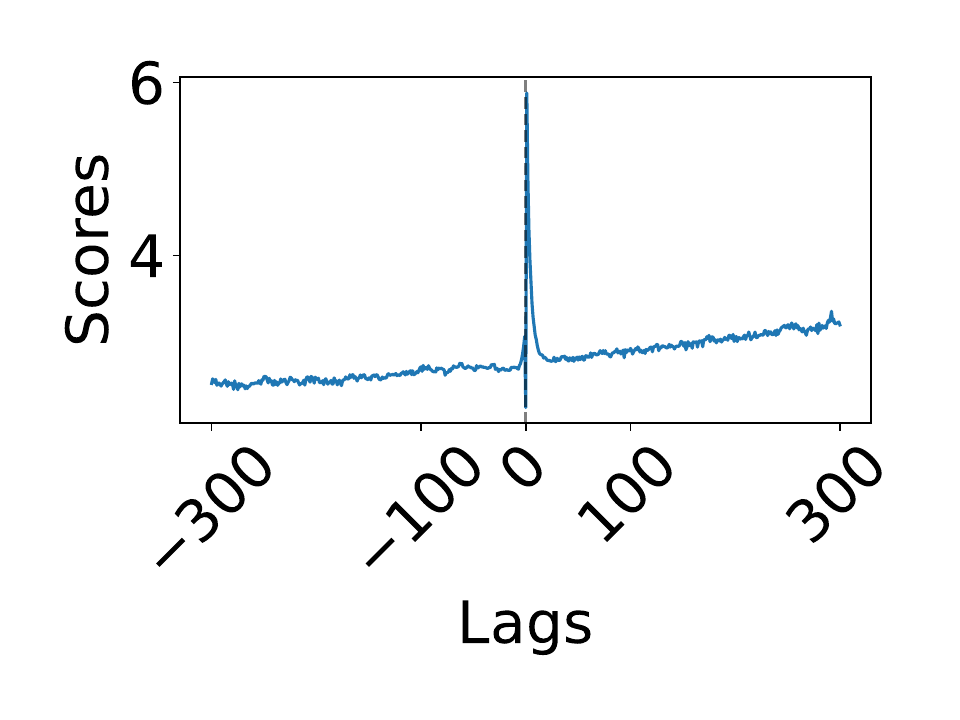}}\\
    \end{tabular}
    \caption{Example of the same induction head L7H3 during different stages of training. \textbf{A.} After 300 iterations. \textbf{B.} After 4000 iterations. \label{fig:crp_training}}
\end{figure}

\begin{figure*}[h!]
    \centering
    \begin{tabular}{lllll}
    \textbf{A} &
    \textbf{B} &
    \textbf{C} &
    \textbf{D} &
    \textbf{E} \\
        {\includegraphics[width=0.17\textwidth]{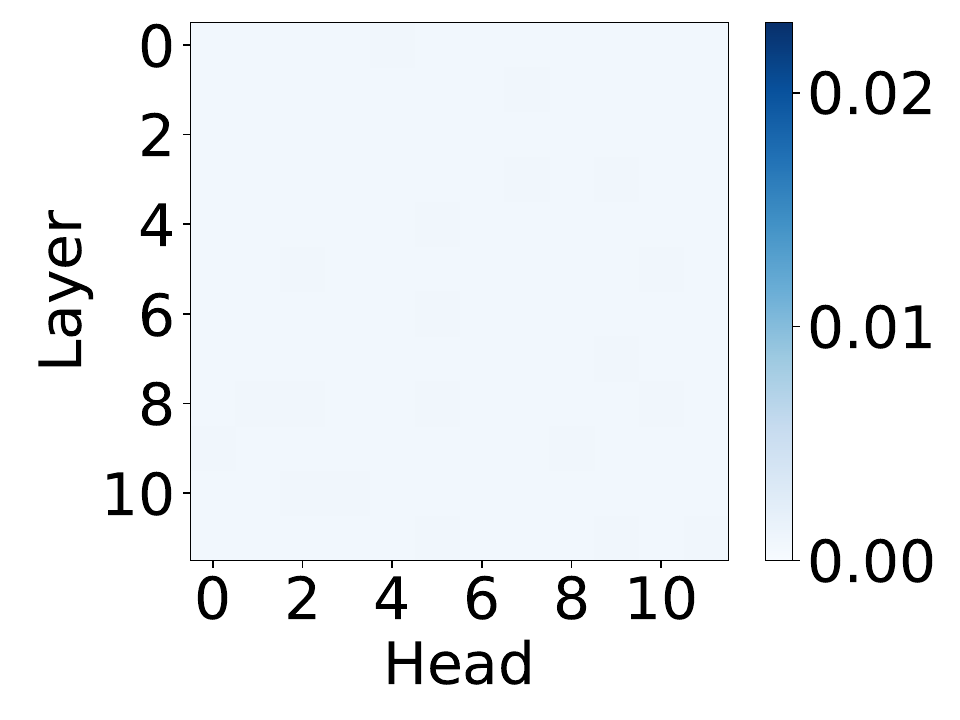}} &
        {\includegraphics[width=0.17\textwidth]{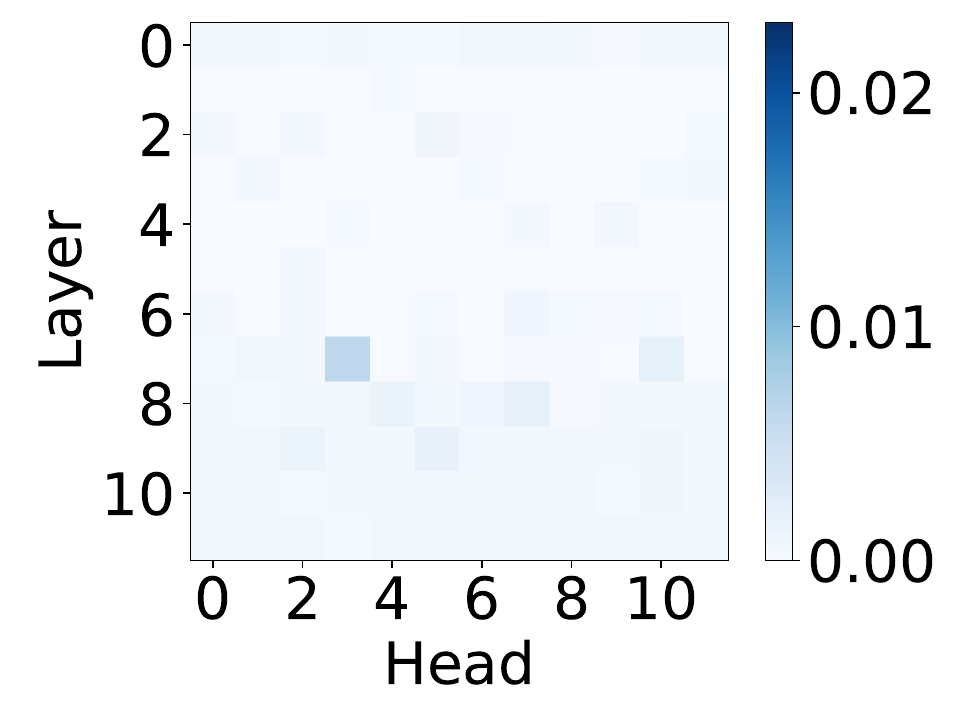}} &
        {\includegraphics[width=0.17\textwidth]{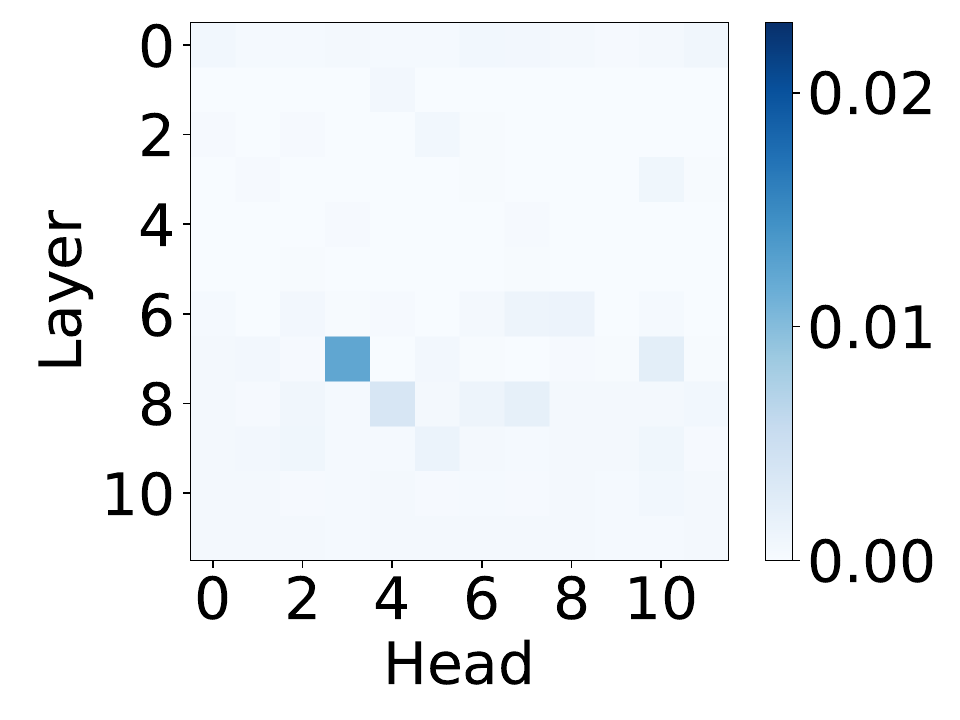}} &
        {\includegraphics[width=0.17\textwidth]{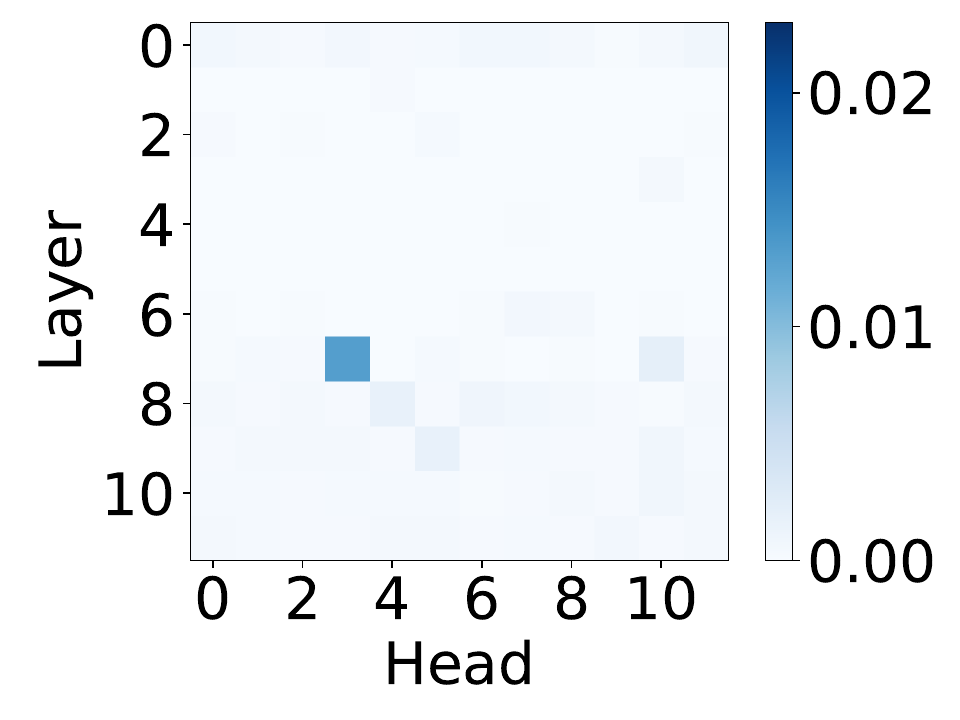}} &
        {\includegraphics[width=0.17\textwidth]{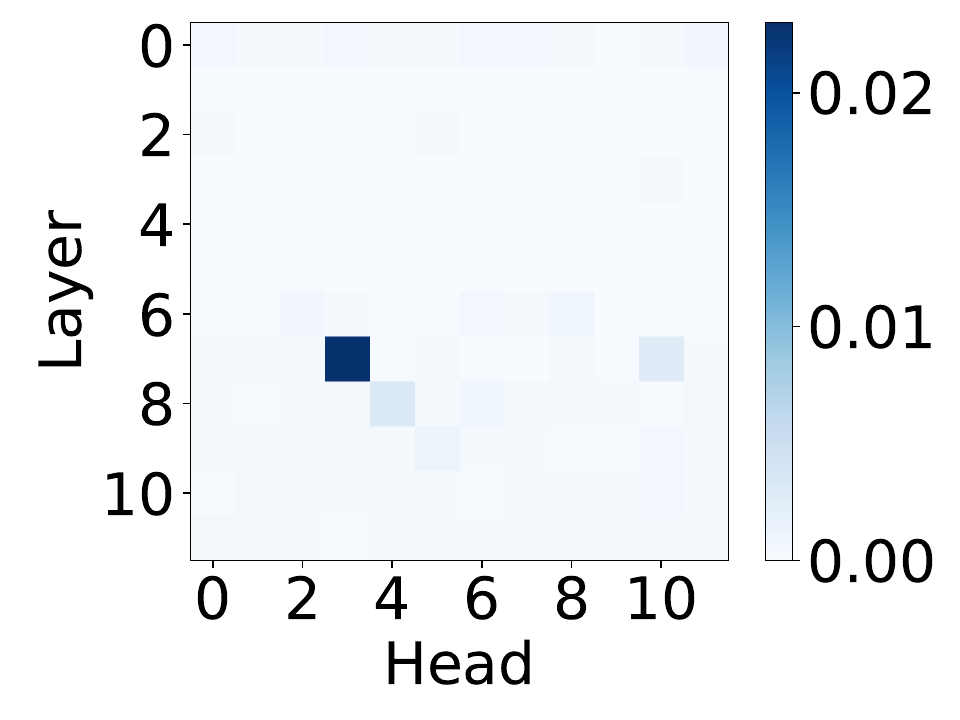}} \\
    \end{tabular}
    \caption{Induction scores for five checkpoints throughout the training of GPT-2 small on Wikitext-103 dataset. \textbf{A.} Random initialization. \textbf{B.} 1000 iterations. \textbf{C.} 2000 iterations.  \textbf{D.} 3000 iterations.  \textbf{E.} 4000 iterations. \label{fig:induction_training}}
\end{figure*}

We observed the increase in the average induction score (averaged across all heads identified as induction heads as defined in the Methods section) and the number of induction heads throughout training (Tab.~\ref{tab:training_iteration_wikitext103} for Wikitext-103, Tab.~\ref{tab:training_iteration_1B} for FineWeb-1B and Tab.~\ref{tab:training_iteration_10B} for FineWeb-10B). For Wikitext-103 we also computed the average recency slope the average time constant fitted to the lag-CRP curve. (These were not computed for FineWeb-1B and FineWeb-10B since their lag-CRP curves for induction heads commonly had non-linear temporal profiles across a wide range of tested lags, making it more difficult to isolate and characterize the recency effect.) 
For Wikitext-103, the average recency slope increased and plateaued around iteration 3000. Prior to training (iteration 0), the average recency slope was also 0. This is because we used trained positional embeddings so prior to training the positional embedding vectors were random. The average time constant fitted to the lag-CRP curve (attention scores as a function of lag) did not systematically change as training progressed but rather converged to a relatively low value around three tokens (note that \textit{lag} is in the units of tokens) after around 3000 iterations.  

To better understand the emergence of induction heads, we visualized the induction scores at all heads at five different training steps on GPT-2 small for Wikitext-103 (Fig.~\ref{fig:induction_training}). Early in training, induction scores are very small in magnitude, and they increase gradually, showing presence mainly in layers six to nine. The locations and number of induction heads did not change during training, showing gradual shaping of the temporal properties. In Fig.~\ref{fig:crp_training}, we showed attention scores as a function of lag for the same head at two different stages of training -- note the order of magnitude change on the y-axis from iteration 300 (Fig.~\ref{fig:crp_training}A) to iteration 4000 (Fig.~\ref{fig:crp_training}B). Fig~\ref{fig:crp_all_GPT2-small_0_1} shows attention scores of all heads before training, Fig~\ref{fig:crp_all_GPT2-small_1000_1} after 1000 iterations and Fig~\ref{fig:crp_all_GPT2-small_4000_1} after 4000 iterations. These figures further illustrate the gradual emergence of temporal profiles, including both induction and recency.

\subsection{Positional encoding magnitude shapes recency and contiguity effects}

\begin{table*}[h!]
\centering
\caption{Impact of positional encoding on temporal properties of attention heads (GPT-2 small trained on Wikitext-103 dataset). }
\label{tab:positional_coding}
\begin{tabular}{@{}lccccc@{}} % 'l' for the metric name column and 'c' for the center-aligned data columns
\toprule
\multicolumn{1}{c}{} & \multicolumn{5}{c}{Positional encoding magnitude} \\
\cmidrule(l){2-6} % Spanning line under the header from column 2 to 6
Metric                       & 0        & 0.5       & 1         & 1.5      & 2               \\ \midrule
Average Induction Score      & 0.0010  & 0.0006   & 0.0024   & 0.0038   & 0.0055            \\
Average Time Constant        & 29.1    & 2.0     & 3.1     & 3.8    & 34.1              \\
Average Recency Slope        & 0    & 0.0034    & 0.0029    & 0.0045   & 0.0053              \\ 
Number of induction heads    & 6    & 25    & 20    & 15   & 13              \\ \bottomrule
\end{tabular}
\end{table*}

\begin{figure}[h!]
    \centering
        {\includegraphics[width=1\columnwidth]{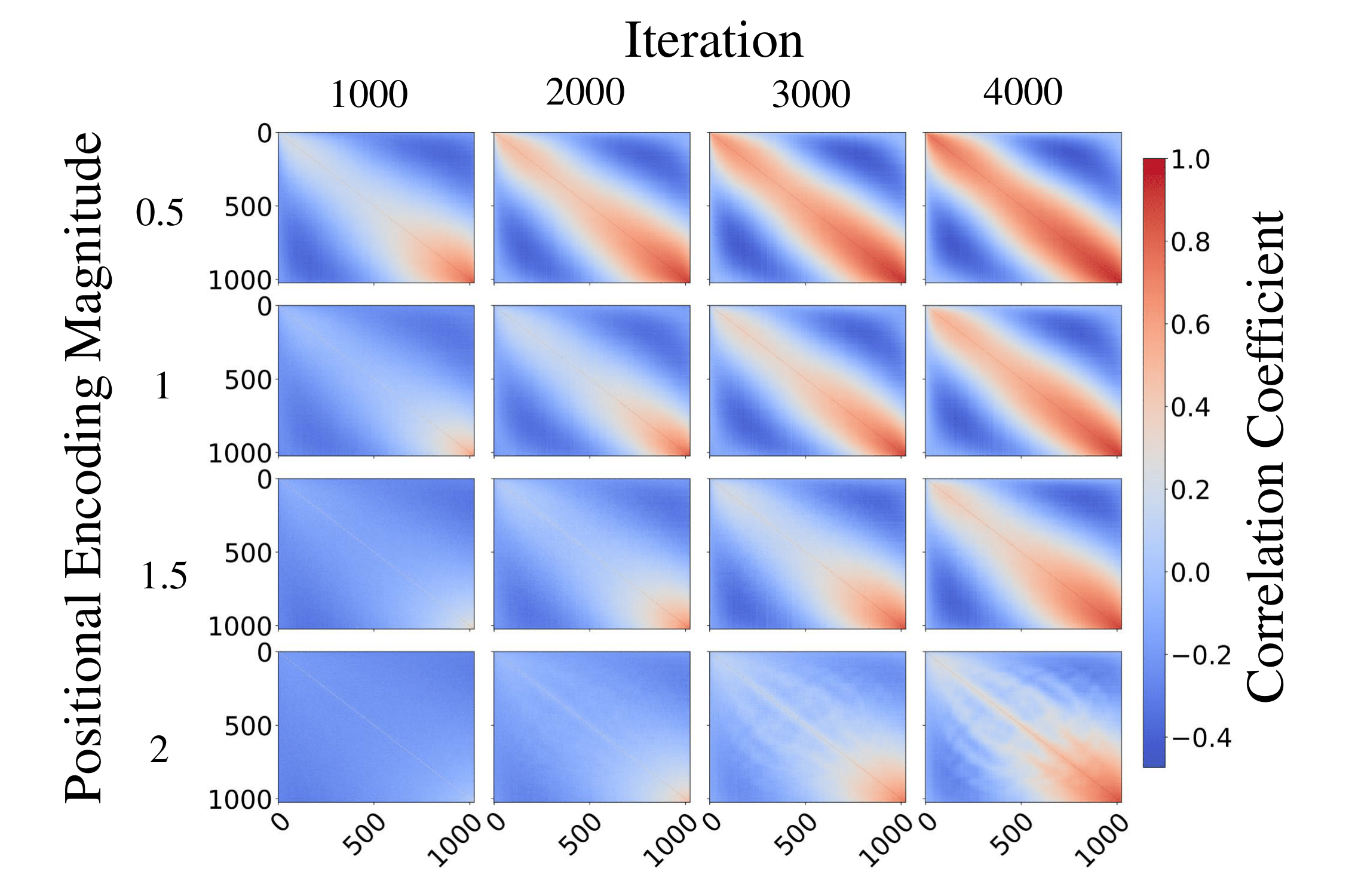}} 
    \caption{Correlation in positional encoding vectors scales with training iterations and positional encoding magnitude during training of GPT-2 small on Wikitext-103 dataset.  \label{fig:learned_positional_coding vectors}}
\end{figure}

\begin{figure*}[h!]
    \centering
    \begin{tabular}{lllll}
    \textbf{A} &
    \textbf{B} &
    \textbf{C} &
    \textbf{D} &
    \textbf{E} \\
        {\includegraphics[width=0.17\textwidth]{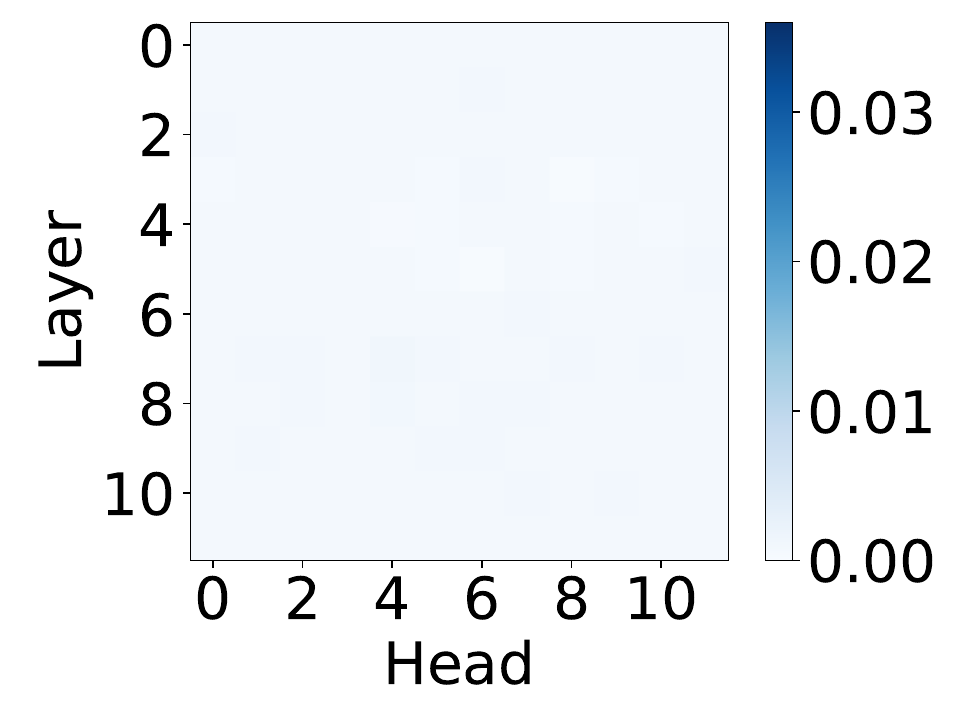}} &
        {\includegraphics[width=0.17\textwidth]{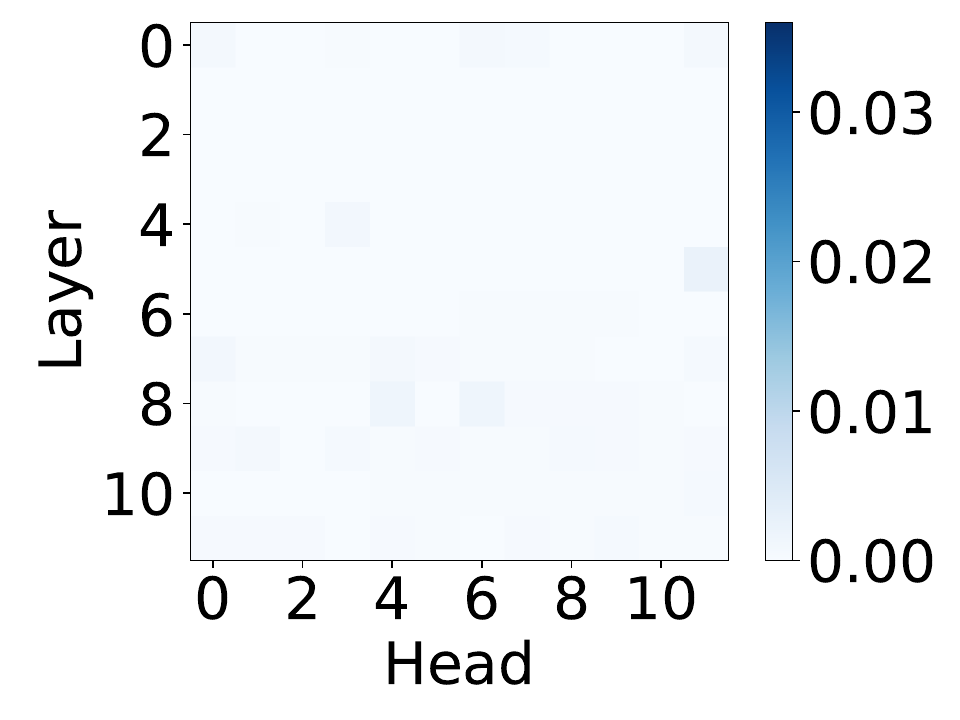}} &
        {\includegraphics[width=0.17\textwidth]{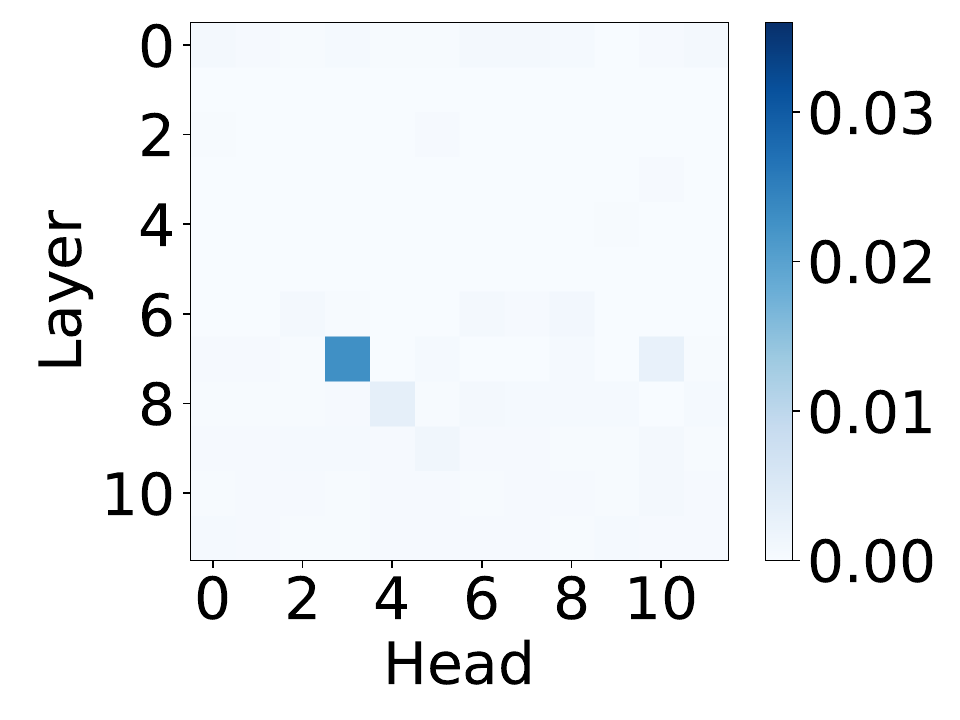}}  &{\includegraphics[width=0.17\textwidth]{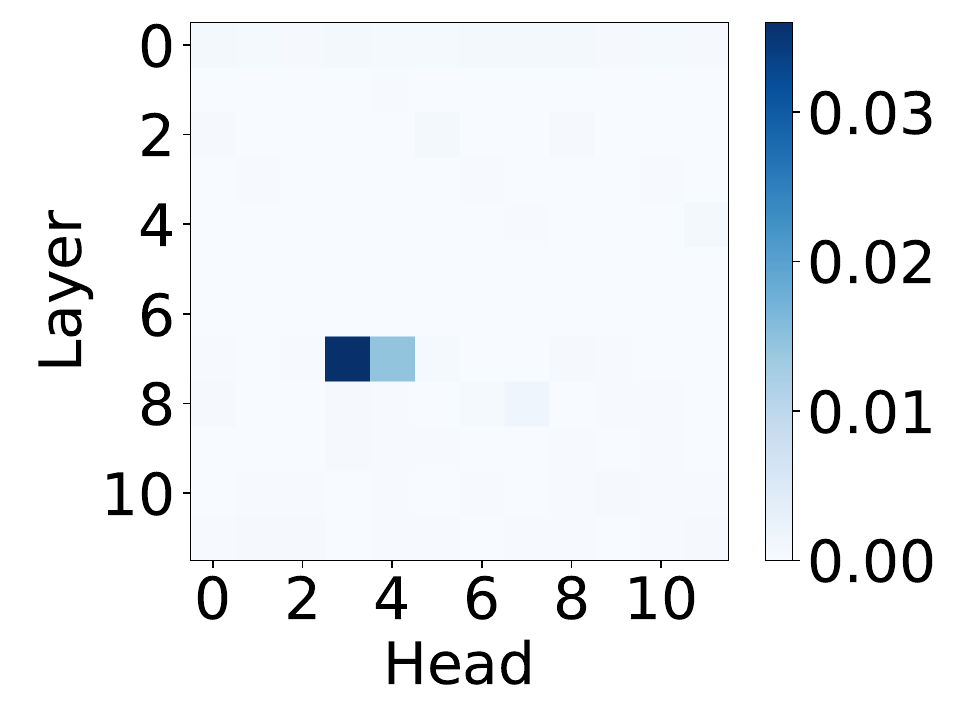}} &
        {\includegraphics[width=0.17\textwidth]{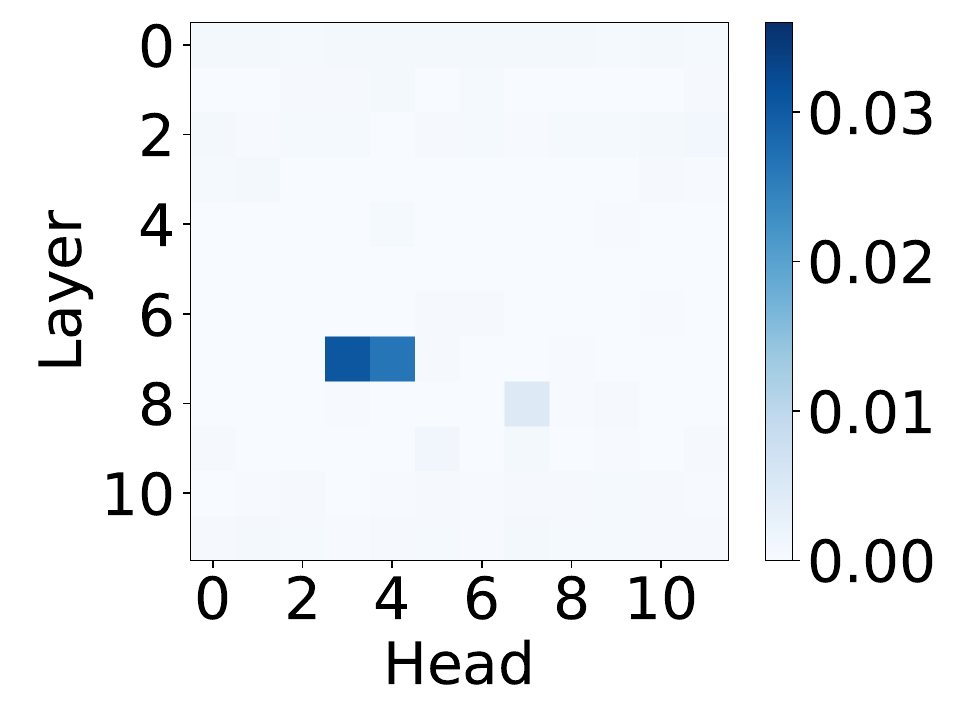}} \\
    \end{tabular}
    \caption{Induction scores for five different magnitudes of positional encoding (GPT-2 small trained on Wikitext-103 dataset). \textbf{A.} No positional encoding, \textbf{B.} 0.5,  \textbf{C.} 1,  \textbf{D.} 1.5,  \textbf{E.} 2. \label{fig:induction_positional_coding}}
\end{figure*}

Increasing the magnitude of positional encoding increased the average induction score (Tab.~\ref{tab:positional_coding}). This is consistent with the hypothesis that vector similarity induced by positional encoding creates a temporal link needed for the induction heads to identify the token that followed the current token in the previous sequence. To further test this hypothesis, we plotted the Pearson correlation coefficient between the positional embeddings for different magnitudes of positional encoding and different training iterations (Fig.~\ref{fig:learned_positional_coding vectors}). The plot reveals an interesting trade-off in the magnitude of positional encoding and training iteration: for example, the correlation profile after 1000 iterations and a magnitude of 0.5 is similar to the correlation profile after 2000 iterations and a magnitude of 1. This relationship is visible across the diagonals of Fig.~\ref{fig:learned_positional_coding vectors}, except for the magnitude of positional encoding equal to 2. For a magnitude of 2, we see more complex temporal patterns that include oscillatory dynamics in the amount of temporal correlations. Overall, the profiles indicate an increase in temporal correlations with training and magnitude of positional encoding, supporting the observed increase in the average induction score shown in Tab.~\ref{tab:positional_coding}. The number of induction heads decreased with the magnitude of positional encoding (Tab.~\ref{tab:positional_coding}). This is also visible in Fig.~\ref{fig:induction_positional_coding}, where the scores of some induction heads increase, while for others, the scores decrease, so they no longer fit the criteria set for induction heads.

\begin{figure*}[h!]
    \centering
    \begin{tabular}{llll}
    \textbf{A} &
    \textbf{B} &
    \textbf{C} &
    \textbf{D}\\
        {\includegraphics[width=0.22\textwidth]{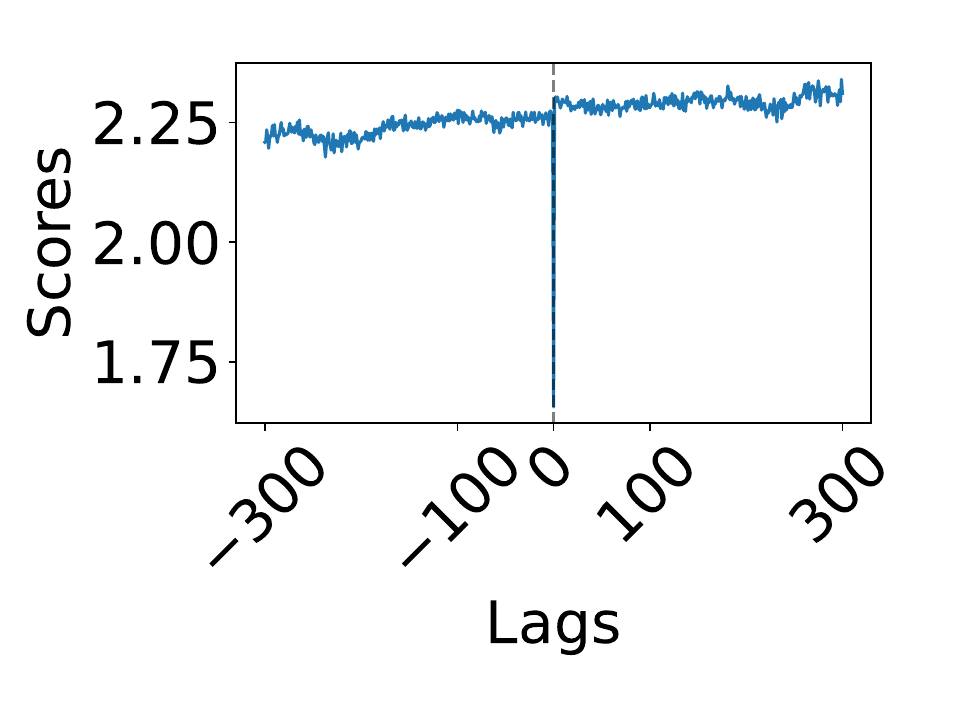}} &
        {\includegraphics[width=0.22\textwidth]{gpt2-baseline-L7H3_4000.pdf}} &
        {\includegraphics[width=0.22\textwidth]{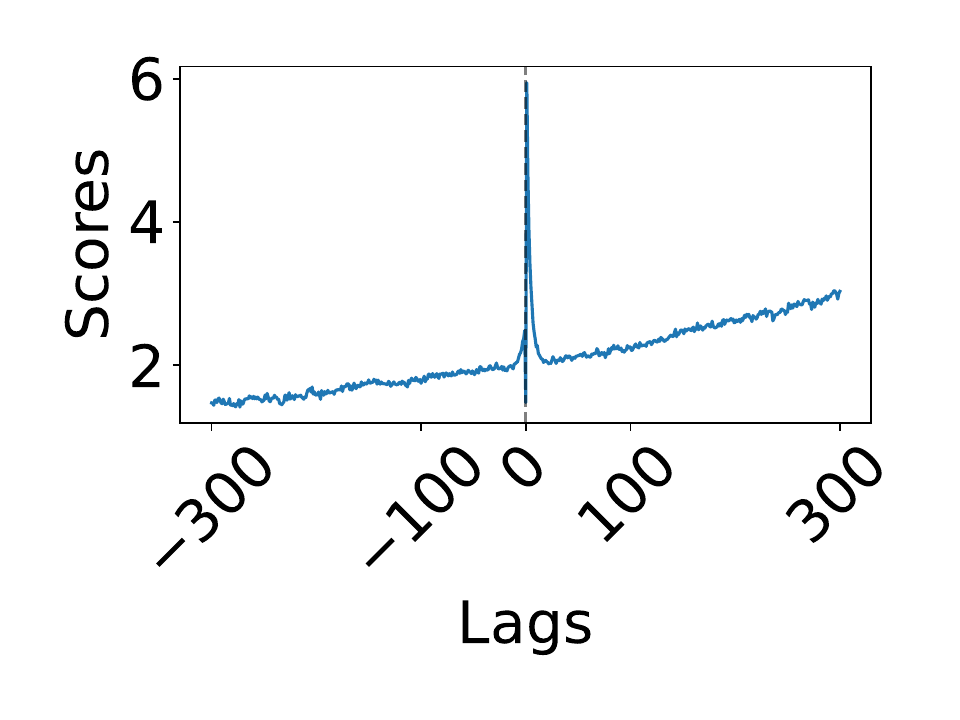}} &{\includegraphics[width=0.22\textwidth]{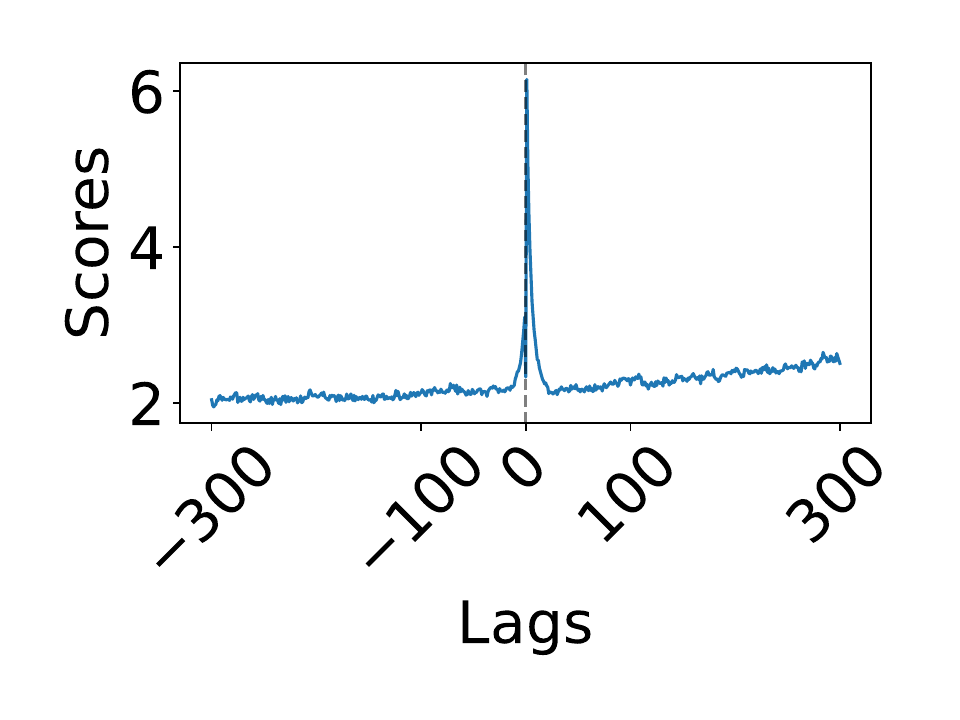}}\\
    \end{tabular}
    \caption{Example of the same induction head L7H3 for different magnitudes of positional encoding. \textbf{A.} No positional encoding. \textbf{B.} Positional encoding with magnitude 1 (baseline model). \textbf{C.} Positional encoding with magnitude 1.5. \textbf{D.} Positional encoding with magnitude 2. \label{fig:crp_positional_coding}}
\end{figure*}

The impact of the magnitude of the positional encoding on the average time constant was mixed. Some heads showed long time constants, including heads in models without positional encoding and in models with double the amount of positional encoding. However, heads with a magnitude of positional encoding equal to 0.5, 1 and 1.5 all had relatively short time constants in the range of 2 to 4 lags. Thus models with these balanced magnitudes of positional encoding did not retrieve extended temporal context. Fig~\ref{fig:crp_positional_coding} shows scores of a single attention head for different magnitudes of positional encoding illustrating relatively short time constants. The average recency slope increased with the magnitude of the positional encoding, as expected, due to increased temporal similarity induced by the positional encoding.

Tab.~\ref{tab:positional_coding} indicates that with no positional encoding, the slope at the six induction heads was 0. A closer look at all of the attention heads with no positional encoding (Fig.~\ref{fig:crp_all_GPT2-small_4000_0}) reveals several heads that exhibited weak recency effect (note that the range of magnitudes of the heads that show the recency effect was typically small without positional encoding). Previous work has argued that causal masking could have similar effects as positional encoding because it introduces sequential dependencies \cite{haviv2022transformer}. These dependencies can result in the encoding of the input order and could explain the weak recency effect.

\subsection{Temporal effects are consistent across models of different sizes}
\begin{table*}[h!]
\centering
\caption{Impact of model size on temporal properties of attention heads.}
\label{tab:models}
\begin{tabular}{@{}lcc@{}} % 'l' for the metric name column and 'c' for the center-aligned data columns
\toprule
\multicolumn{1}{c}{} & \multicolumn{2}{c}{Model type} \\
\cmidrule(l){2-3} % Spanning line under the header from column 2 to 6
Metric                 & GPT-2 small     & GPT-2 medium                           \\ \midrule
Average Induction Score        & 0.0024        & 0.0012            \\
Average Time Constant     & 3.1         & 12.1                 \\
Average Recency Slope     & 0.003         & 0.007          \\ 
Number of Induction Heads   & 20  (out of 144, 14\%)          & 45 (out of 384, 12\%)          \\ \bottomrule
\end{tabular}
\end{table*}

\begin{figure}[h!]
    \centering
    \begin{tabular}{ll}
    \textbf{A} &
    \textbf{B} \\
        {\includegraphics[width=0.21\textwidth]{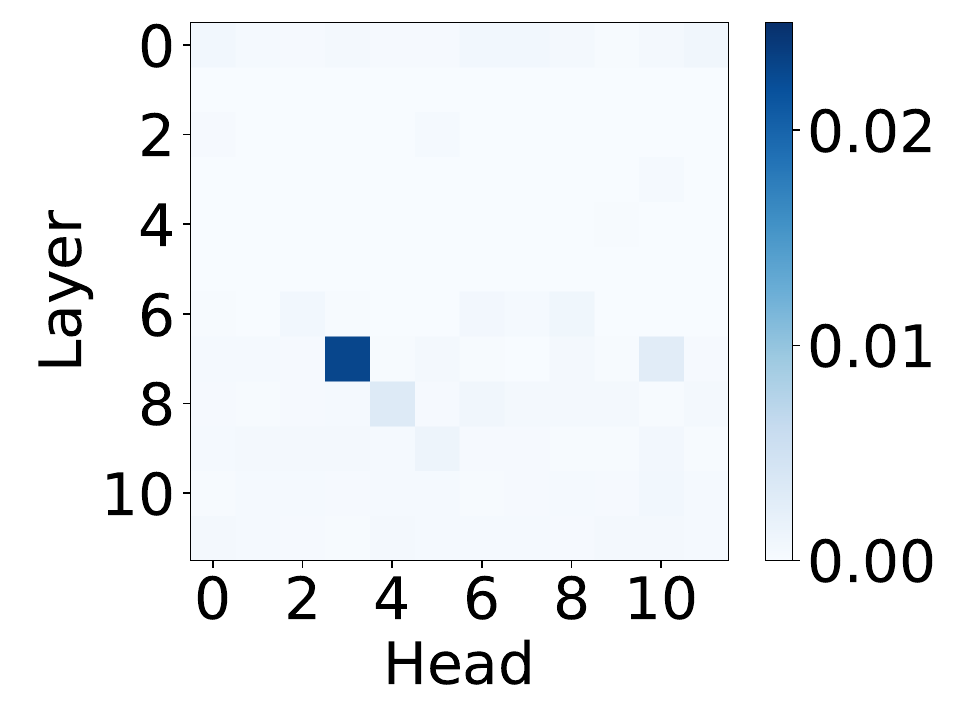}} &
        {\includegraphics[width=0.21\textwidth]{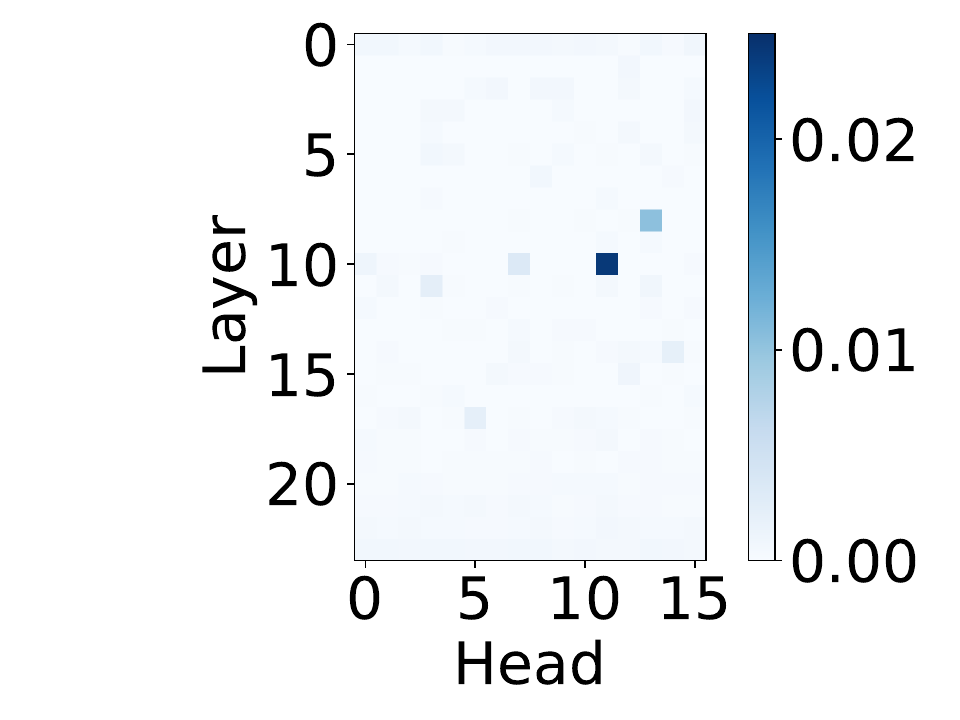}} \\
    \end{tabular}
    \caption{Induction scores for two different models. \textbf{A.} GPT-2 small, \textbf{B.} GPT-2 medium. \label{fig:induction_models}}
\end{figure}

All previous results were discussed for GPT-2 small model. We also trained GPT-2 medium but only for the baseline magnitude of positional encoding and on Wikitext-103. Overall, we observed similar attention score profiles across the two models (compare Fig.~\ref{fig:crp_all_GPT2-small_4000_1} and Fig.~\ref{fig:crp_all_GPT-medium_2000_1}). We quantified these observations in Tab.~\ref{tab:models}. Fig~\ref{fig:induction_models} shows induction scores in the two models, indicating that GPT-2 medium had larger scores concentrated closer to the input layer than GPT-2 small.

\begin{figure}[h!]
    \centering
    \includegraphics[width=1\columnwidth]{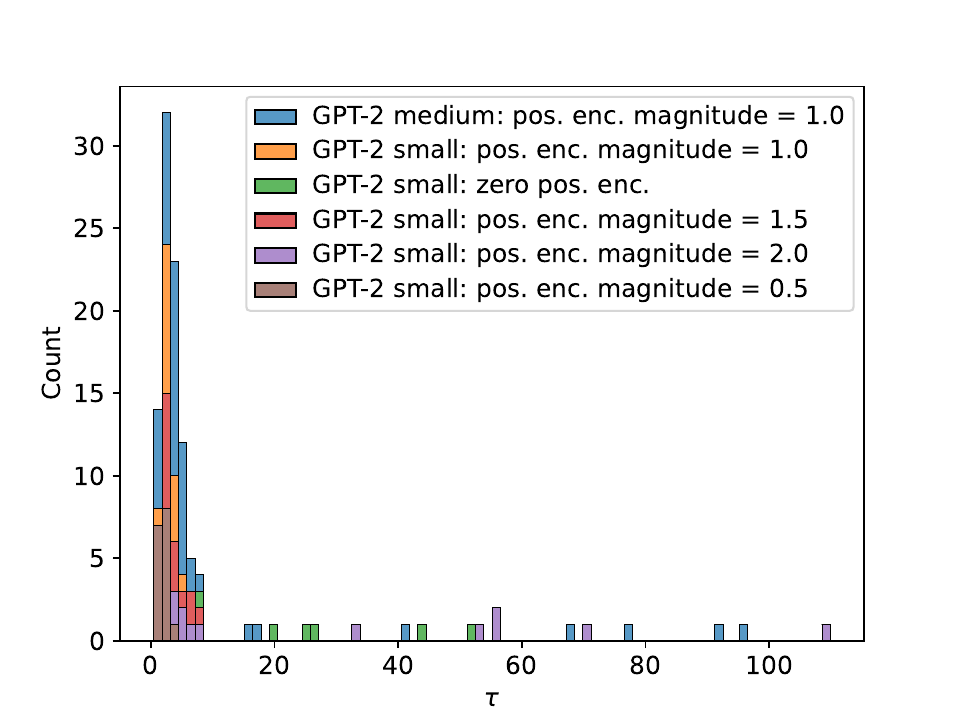}
    \caption{Distributions of fitted time constants of induction heads for different models and magnitudes of positional encoding.}
    \label{fig:tau-distribution}
\end{figure}

\begin{figure*}[h!]
    \centering
    \begin{tabular}{lll}
    \textbf{A} &
    \textbf{B} &
    \textbf{C} \\
        {\includegraphics[width=0.30\textwidth]{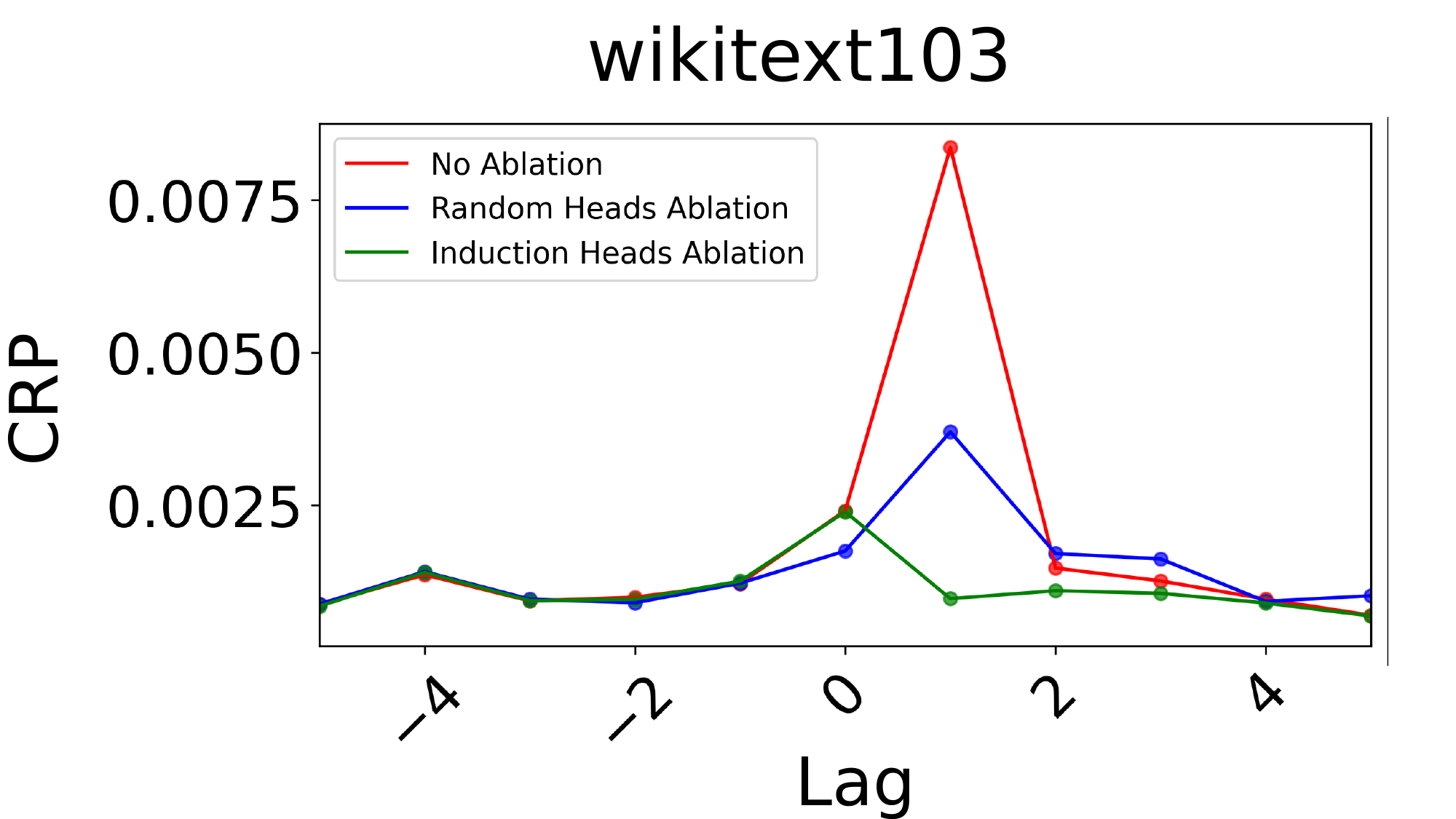}} &
        {\includegraphics[width=0.30\textwidth]{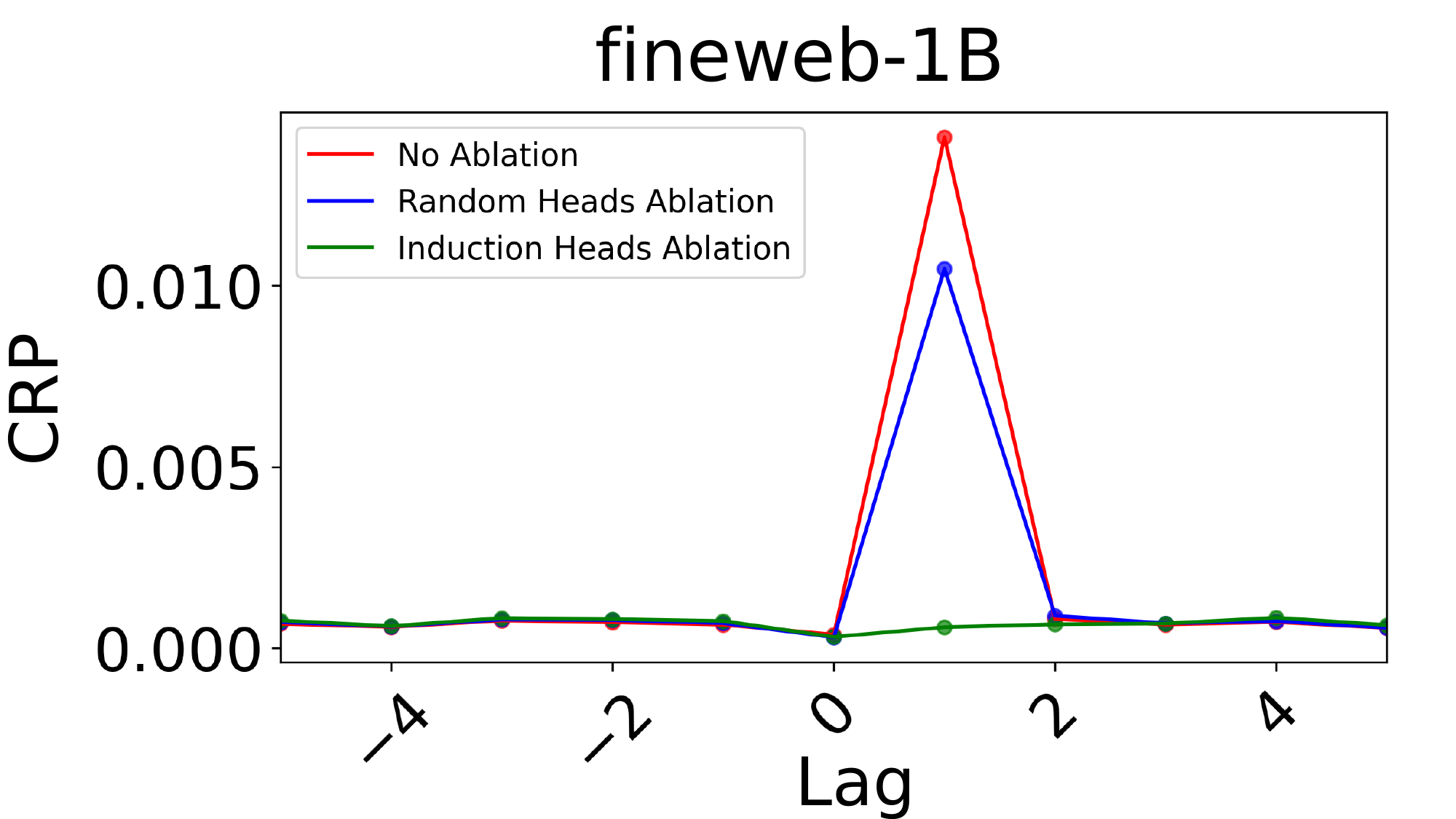}} &
        {\includegraphics[width=0.30\textwidth]{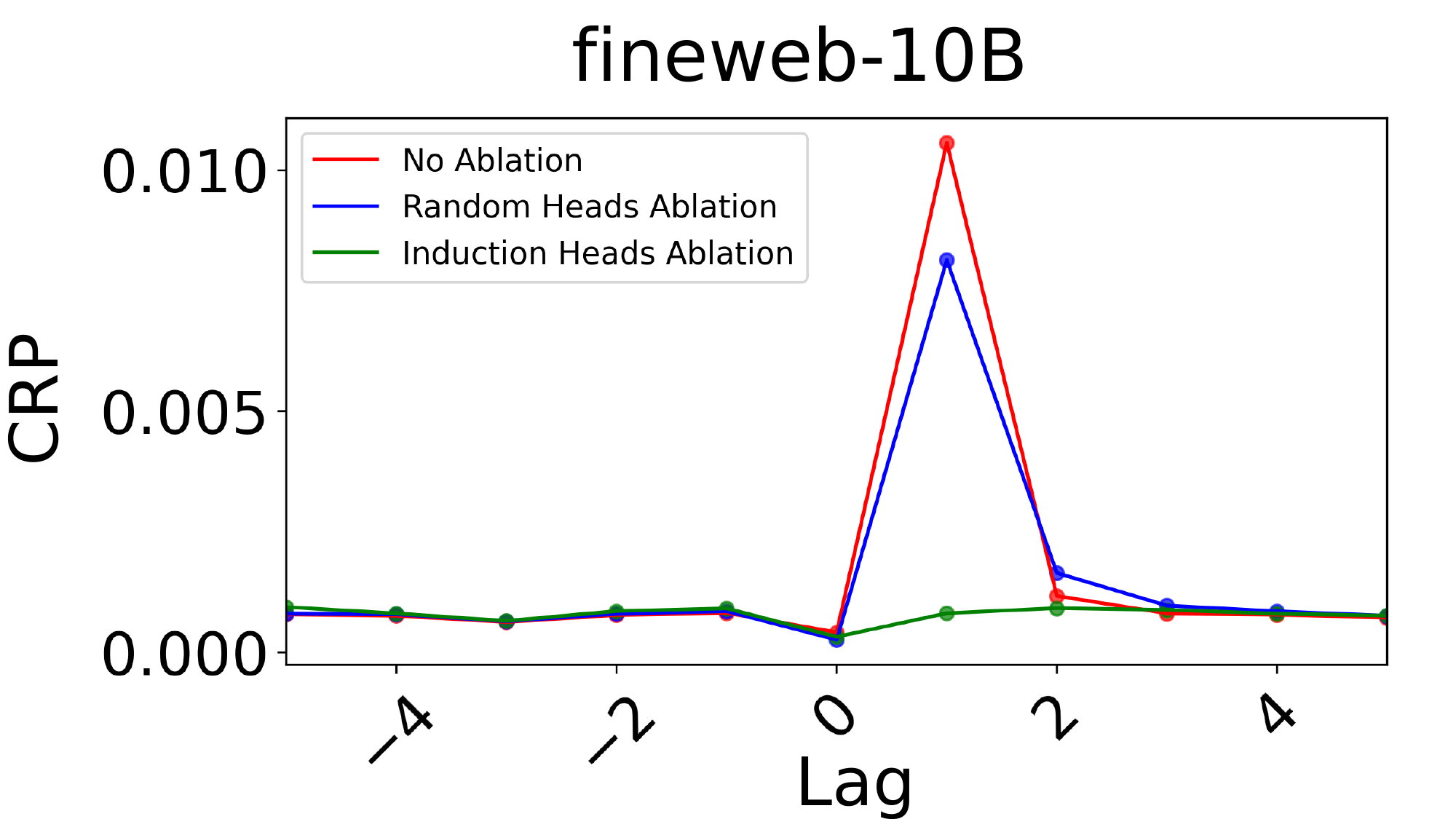}} \\
    \end{tabular}
    \caption{CRP during downstream evaluation showing impact of induction head ablation. \label{fig:crp_downstream_abl_5}}
\end{figure*}

\subsection{Characterizing the contiguity effect across the attention heads}

To better understand the span of temporal context retrieval in transformers, we investigated the distribution of the time constants from fitting the attention scores as a function of lag. We found that the time constants are mainly concentrated in the narrow range of 2-4 lags, with only a few heads covering larger lags. This result holds for models with different magnitudes of positional encoding and for both model sizes. This suggests that when retrieving in-context information given a repeated token, aside from primacy and recency effects, transformers will primarily focus attention on the very local (2-4 lags) neighborhood of that token.

\begin{figure}[h!]
    \centering
        {\includegraphics[width=1\columnwidth]{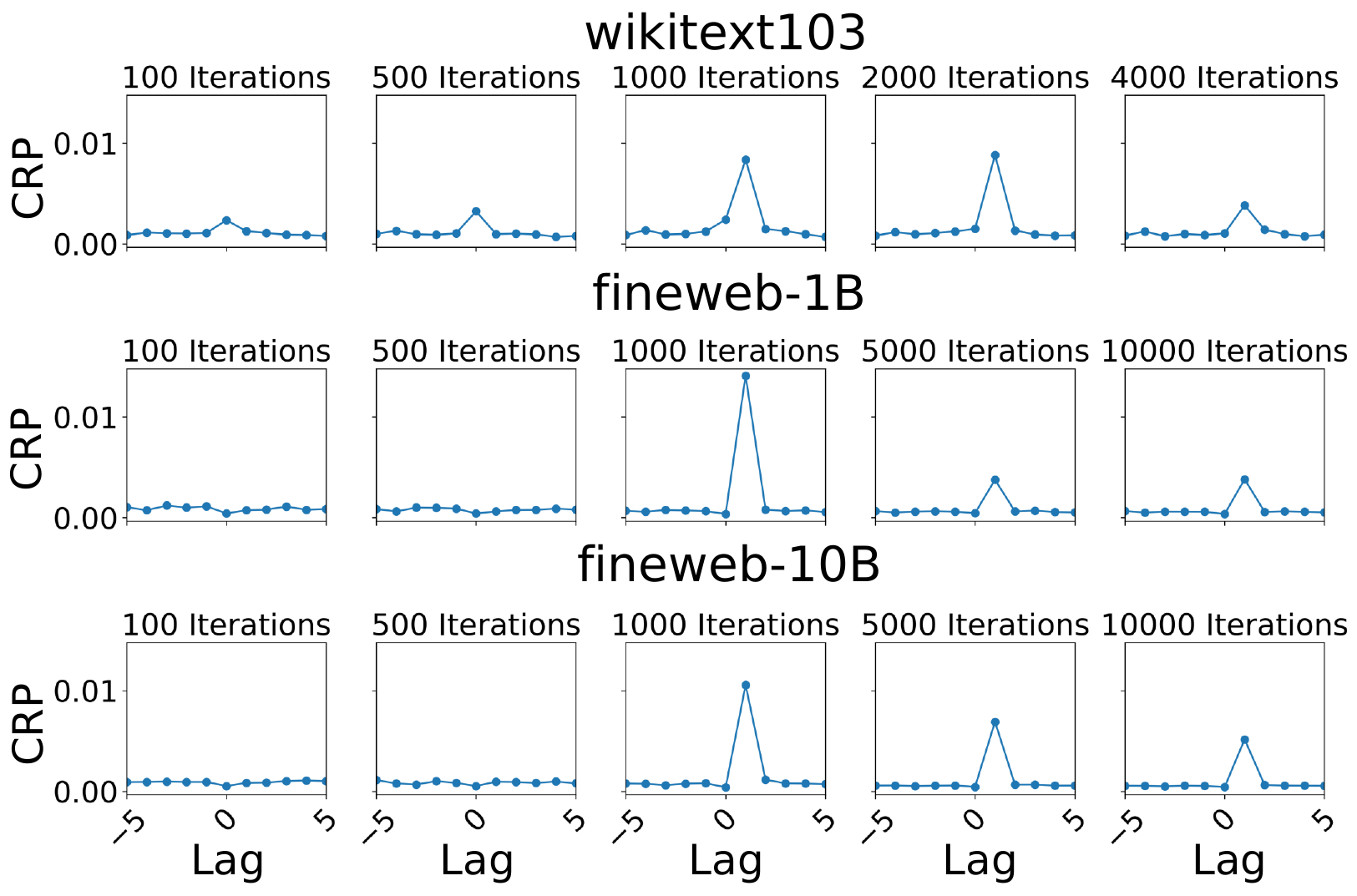}} 
    \caption{CRP as a function of training iteration in downstream evaluation.  \label{fig:crp_downstream_iter_5}}
\end{figure}

\subsection{Downstream evaluation and impact of induction heads ablation}

To better understand the impact of temporal context retrieval in attention heads on transformer outputs, we conducted a downstream evaluation inspired by the free recall memory task. 
After training the models, we probed them with a sequence of 500 randomly ordered tokens (we selected 500 tokens that were most frequently occurring in each dataset) followed by a middle token (e.g., \textit{GRDBTHMB}, where each character corresponds to an individual token). We then quantified the probability of the next token as a function of lag (distance from the middle token). Temporal contiguity predicts a larger probability for tokens that are temporally adjacent to the middle token, while recency and primacy effects predict a higher probability for tokens from the beginning (large negative lag) and end (large positive lag) of the list. 

During the training (after about 1000 iterations), we observed a strong (around 10 times) increase in the probability of recall for items at lag +1 (Fig.~\ref{fig:crp_downstream_iter_5}, Fig.~\ref{fig:crp_downstream_iter_250}), indicating strong preference for serial recall. To investigate the relationship between induction heads and this effect, we ablated the heads that had induction scores above 0.01 (the ablation was done similarly to \citet{crosbie2024induction} by setting the attention scores for ablated heads to $-\infty$). Even though the number of ablated heads was around 5\% of the total number of heads, the ablation of induction heads eliminated the contiguity effect (Fig.~\ref{fig:crp_downstream_abl_5}, Fig.~\ref{fig:crp_downstream_abl_250}). Ablating the same number of non-induction heads in a layer-matched fashion made much smaller impact on the output probabilities, especially for FineWeb-1B and 10B.

\section{Discussion}
We quantified temporal properties of attention patterns in transformer outputs using lag-CRP analysis, commonly used for studying episodic memory and serial position effects in human memory experiments. 
By using multiple permutations of the input sequences, we were able to reduce the semantic effects of token similarity and isolate the temporal effects making it possible to observe primacy, recency and contiguity effects in the attention heads. 

Unlike human memory experiments, where the contiguity effect is robust across a wide range of scales  \cite{howard2008persistence}, supporting power-law decay of memory \cite{wixted1991form,rubin1996one,donkin2012power}, we did not find evidence for retrieval of a broad temporal context in transformers. In fact, training typically had an impact of reducing the time constants of lag-CRP to small values in the range of 2 to 4 lags. Downstream analysis demonstrated a strong preference towards a serial recall that was eliminated after ablation of the induction heads. Overall, we showed that tools from cognitive science can be used to better understand learning in transformers, providing valuable insights into the emergence of temporal structure during in-context learning.

\section{Limitations}
Our approach used relatively small models. Training larger, instruction fine-tuned language models could provide additional insights into the temporal properties of in-context learning. While our approach was inspired by studies of episodic memory in humans, a number of methodological differences between the present analysis and human experiments (such as the fact that humans receive task instructions) prevent us from making direct parallels with human memory and learning. 

\section*{Acknowledgment}
This research was supported in part by Lilly Endowment, Inc., through its support for the Indiana University Pervasive Technology Institute.

\bibliographystyle{acl_natbib}

\appendix

\newpage

\section*{Appendix}

\label{sec:appendix} 
Below we provide plots showing CRP for downstream evaluation, attention scores as a function of lag for all transformer heads for different models (positional encoding magnitude and model size) and different training stages.

\renewcommand{\thetable}{A\arabic{table}}
\renewcommand{\thefigure}{A\arabic{figure}}
\setcounter{table}{0}
\setcounter{figure}{0}

\begin{figure*}[h!]
    \centering
        {\includegraphics[width=0.5\textwidth]{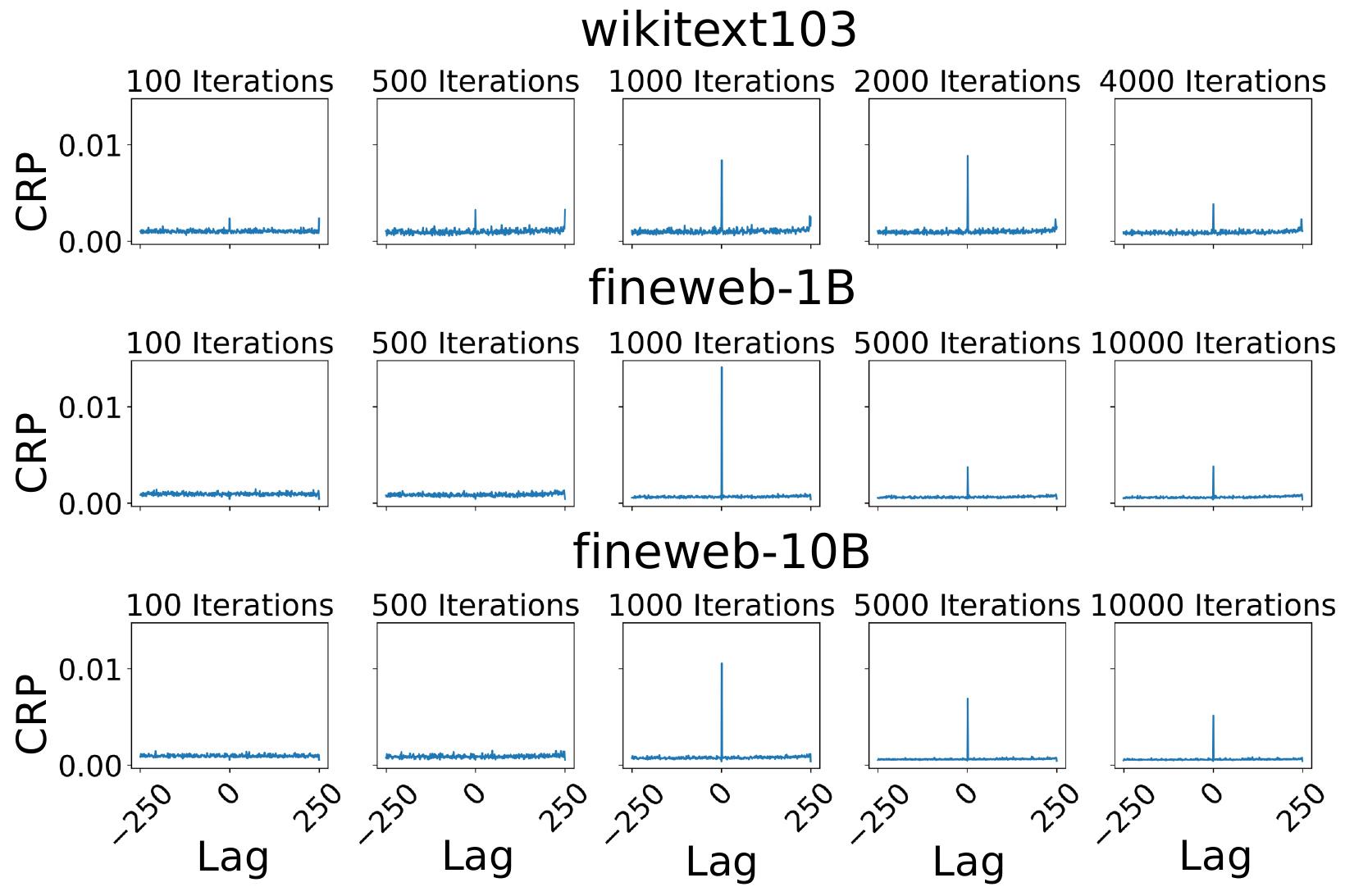}} 
    \caption{CRP as a function of training iteration in downstream evaluation (same as Fig.~\ref{fig:crp_downstream_iter_5} but for more lags).  \label{fig:crp_downstream_iter_250}}
\end{figure*}

\begin{figure*}[h!]
    \centering
    \begin{tabular}{lll}
    \textbf{A} &
    \textbf{B} &
    \textbf{C} \\
        {\includegraphics[width=0.30\textwidth]{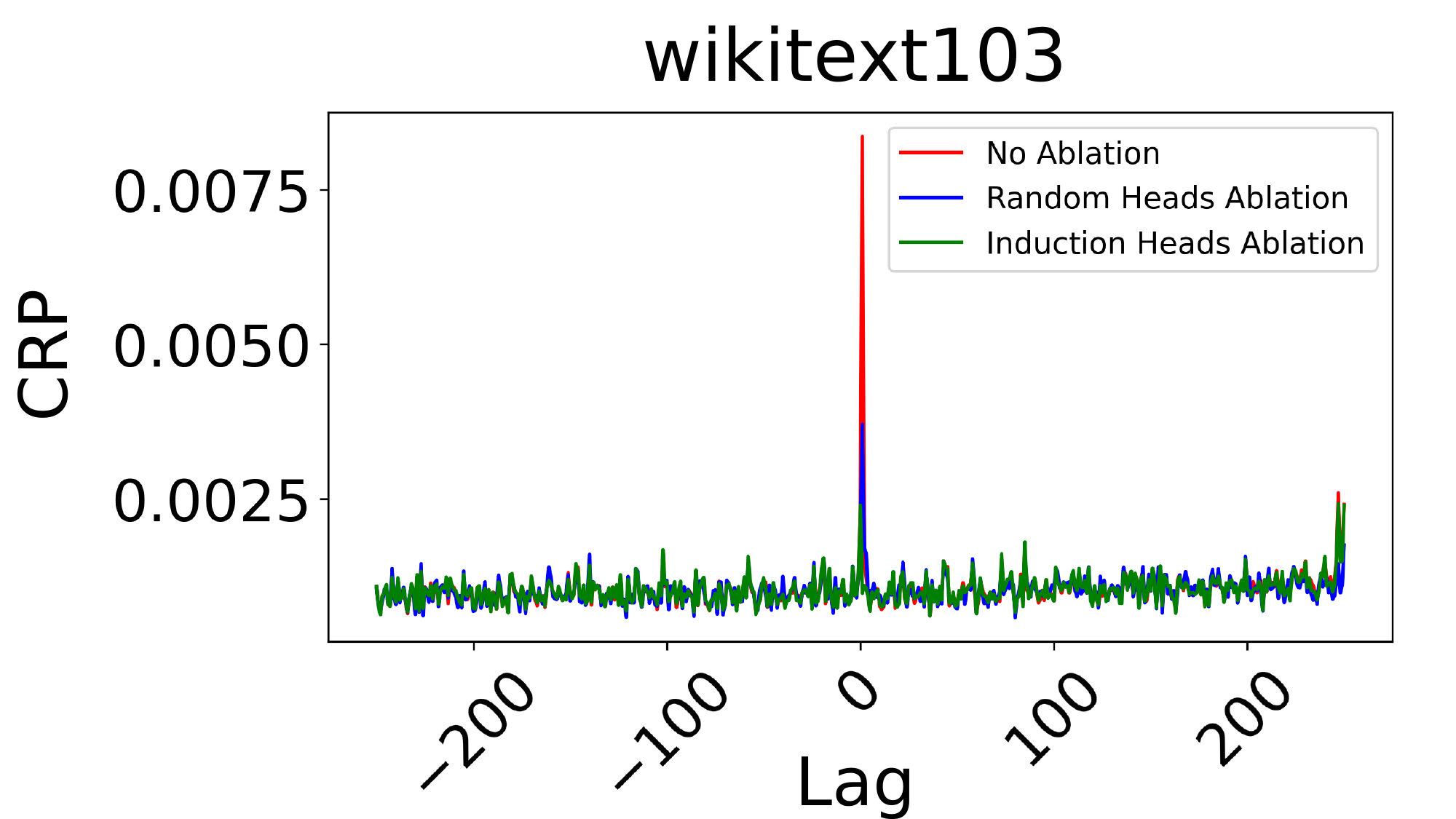}} &
        {\includegraphics[width=0.30\textwidth]{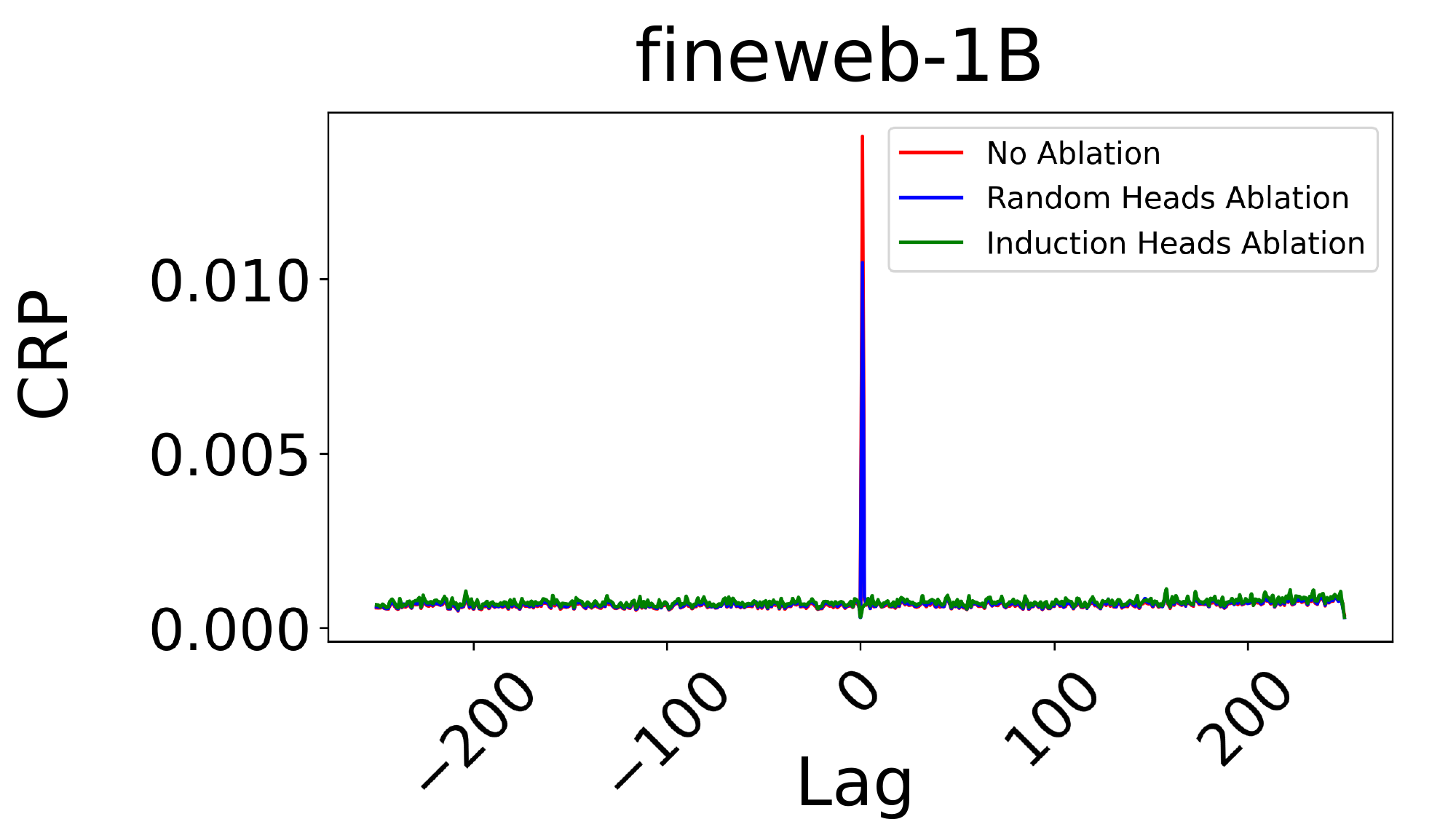}} &
        {\includegraphics[width=0.30\textwidth]{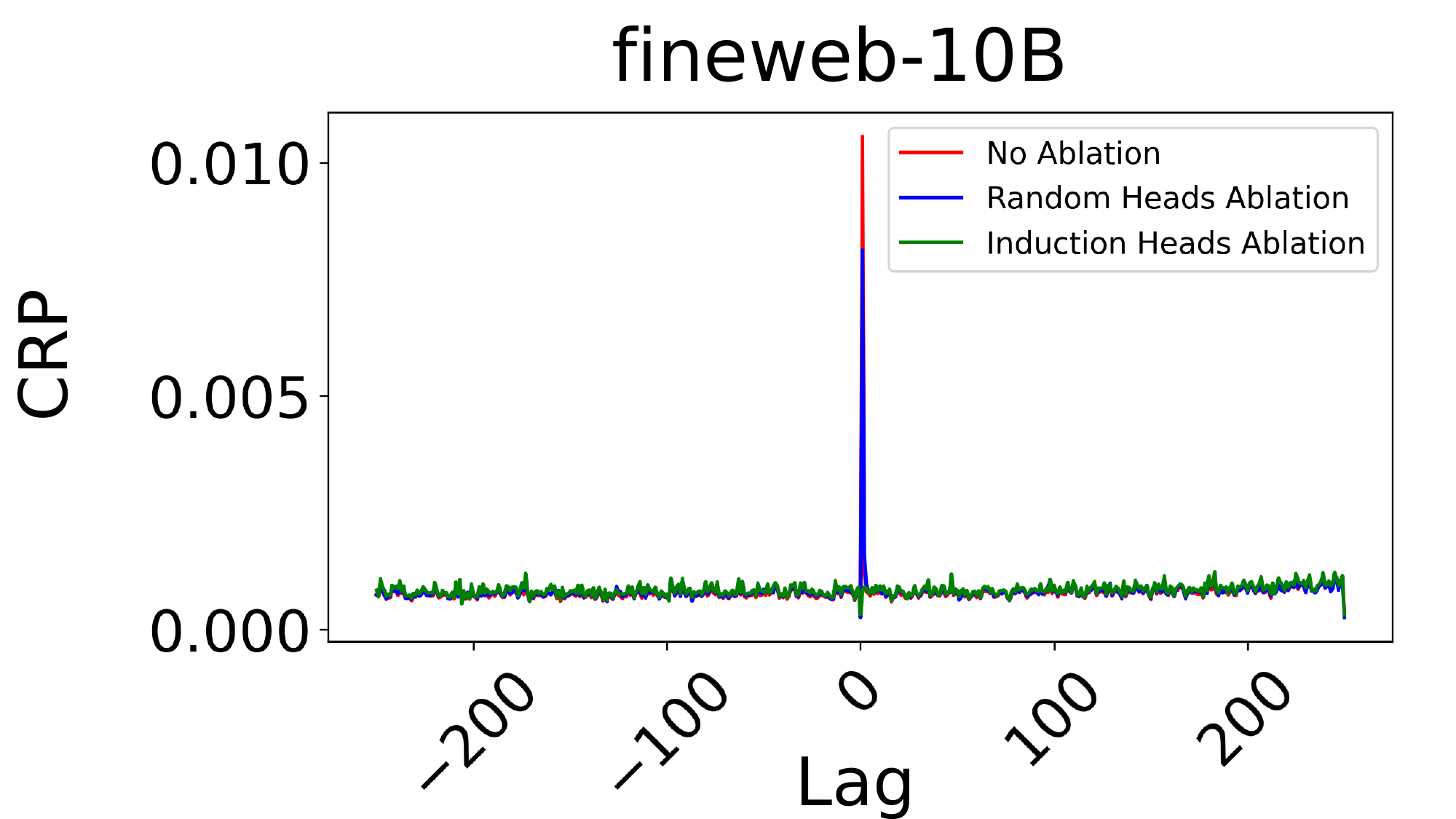}} \\
    \end{tabular}
    \caption{CRP during downstream evaluation showing impact of induction head ablation (same as Fig.~\ref{fig:crp_downstream_abl_5} but for more lags). \label{fig:crp_downstream_abl_250}}
\end{figure*}

\newpage

\begin{figure*}[h!]
    \centering
    \includegraphics[width=1\textwidth]{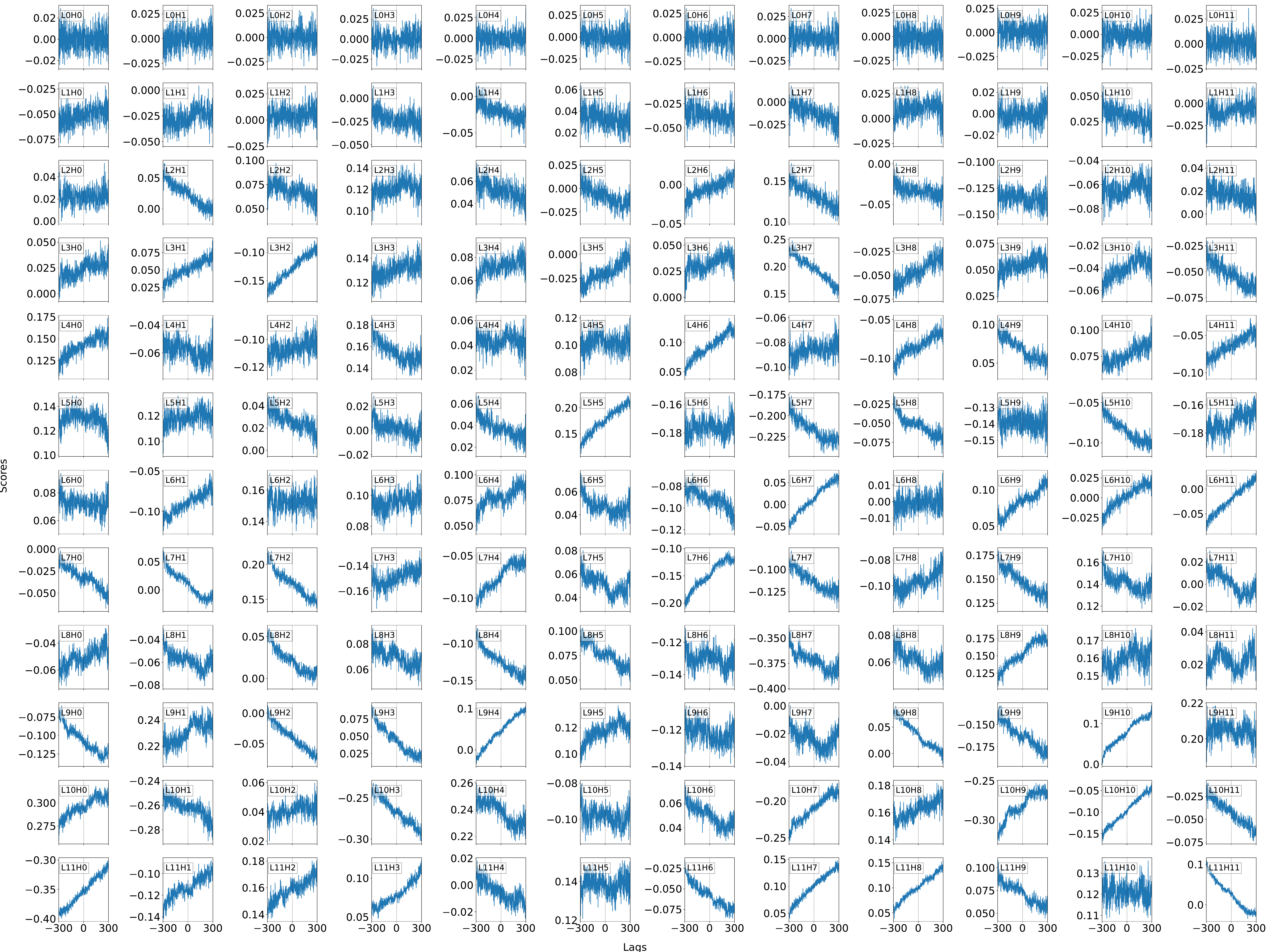}
    \caption{Attention scores as a function of lag for all heads of GPT-2 small before training. \label{fig:crp_all_GPT2-small_0_1}} 
\end{figure*}

\begin{figure*}[h!]
    \centering
    \includegraphics[width=1\textwidth]{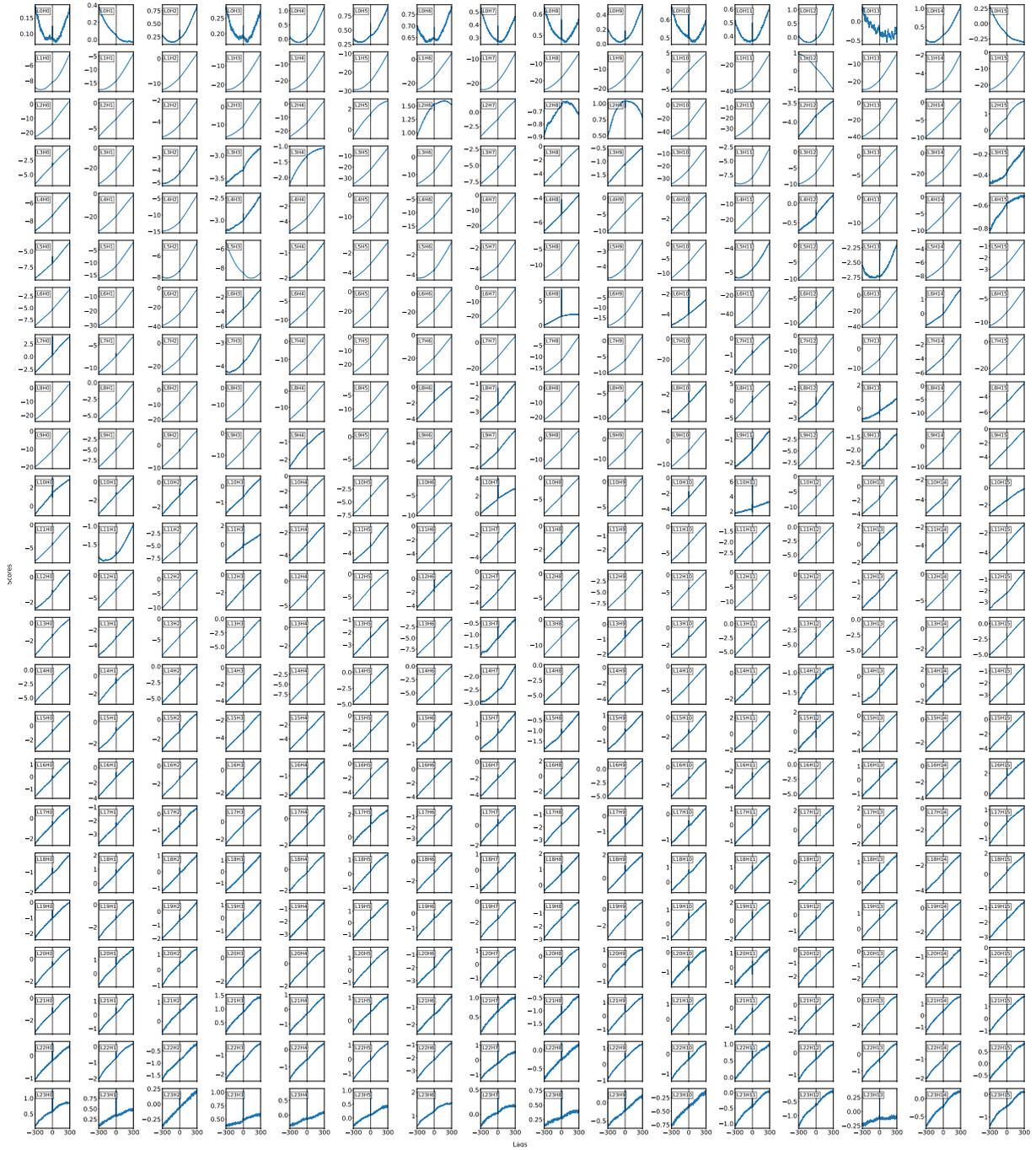}
    \caption{Attention scores as a function of lag for all heads of GPT-2 medium after 2000 iterations for WikiText-103 dataset with a standard amount of positional encoding. \label{fig:crp_all_GPT-medium_2000_1}} 
\end{figure*}

\begin{figure*}[h!]
    \centering
    \includegraphics[width=1\textwidth]{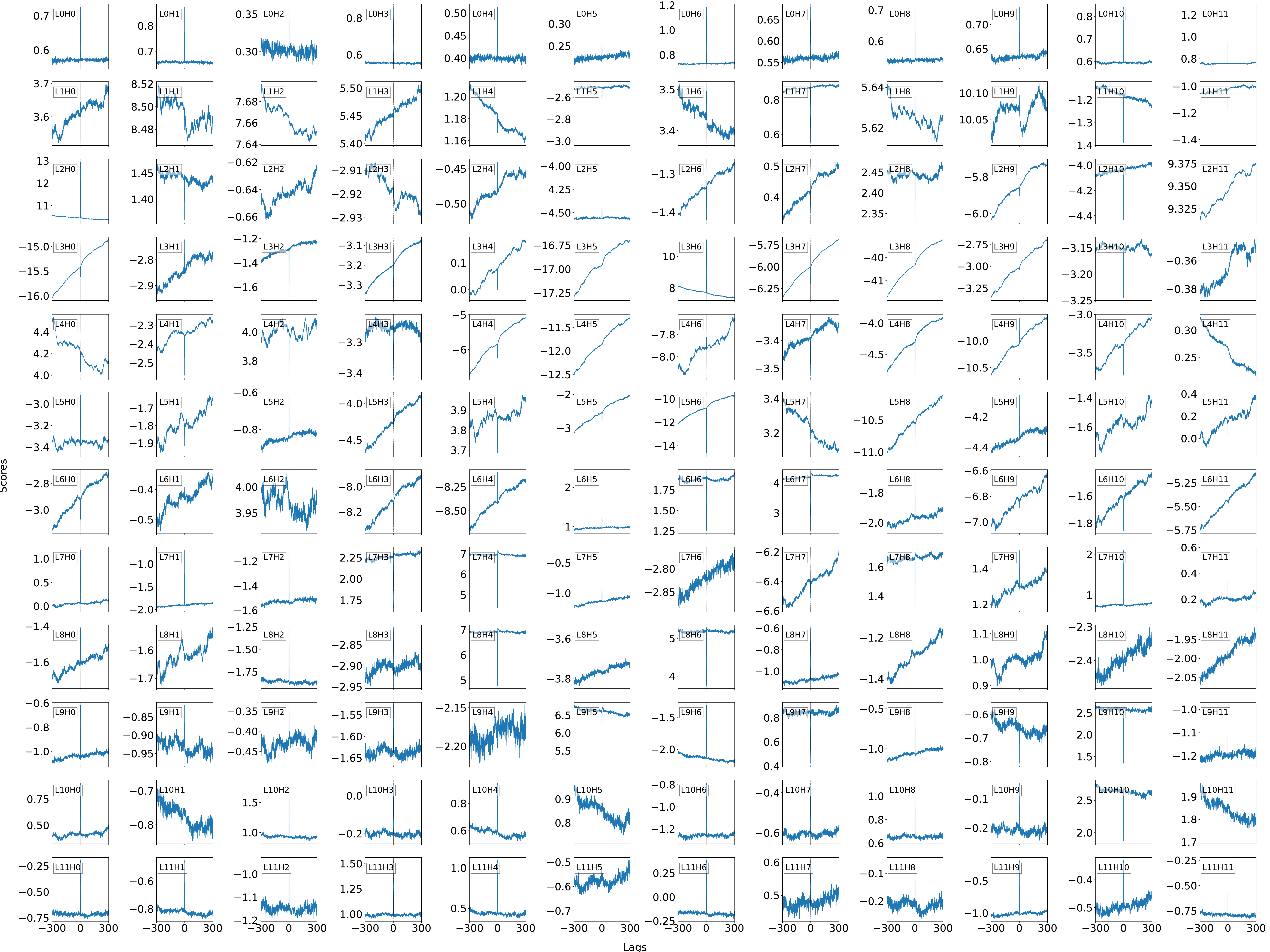}
    \caption{Attention scores as a function of lag for all heads of GPT-2 small after 4000 iterations for WikiText-103 dataset with no positional encoding. \label{fig:crp_all_GPT2-small_4000_0}} 
\end{figure*}

\begin{figure*}[h!]
    \centering
    \includegraphics[width=1\textwidth]{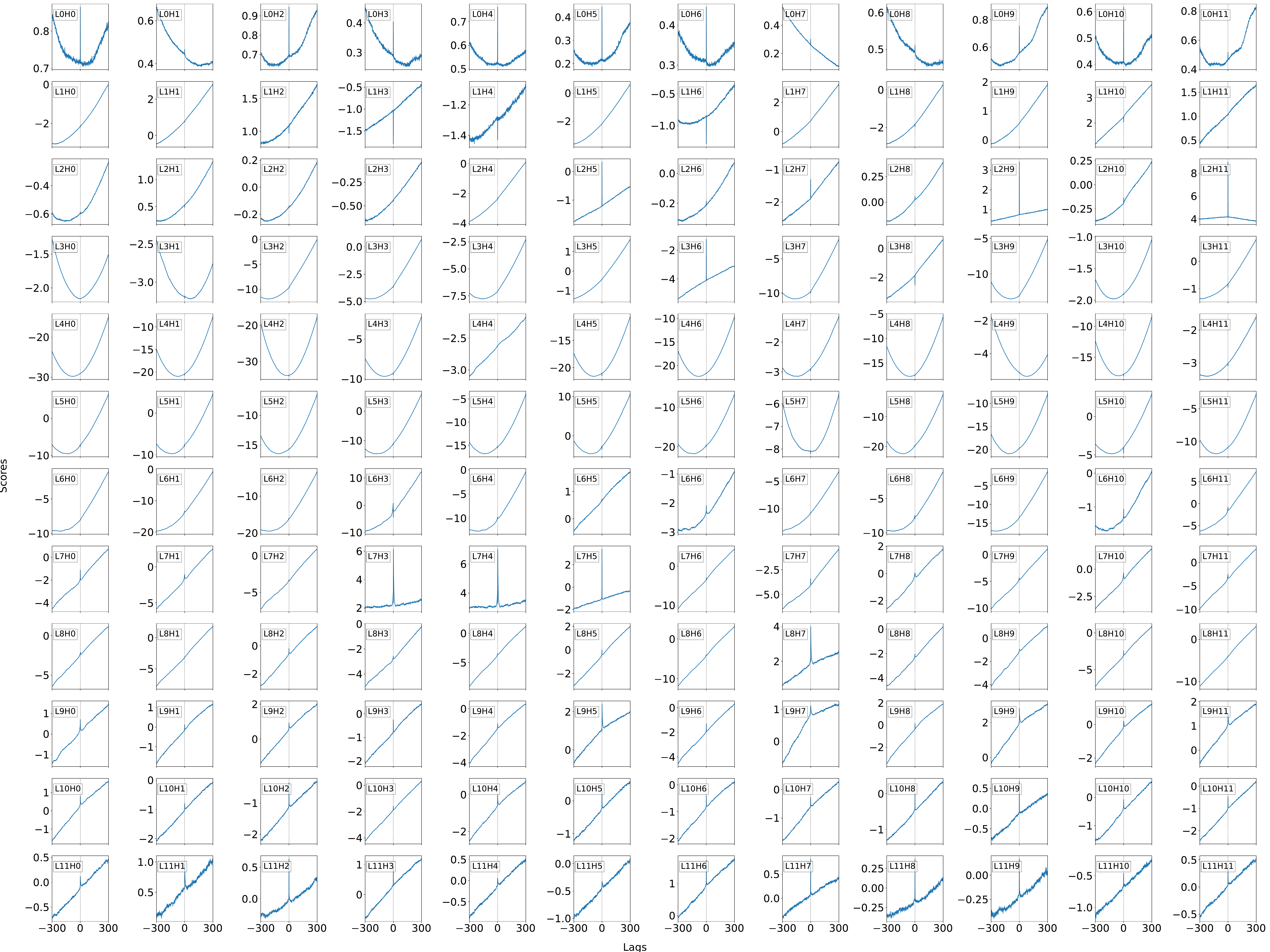}
    \caption{Attention scores as a function of lag for all heads of GPT-2 small after 4000 iterations for Wikitext-103 dataset with double amount of baseline positional encoding. \label{fig:crp_all_GPT2-small_4000_2}} 
\end{figure*}

\begin{figure*}[h!]
    \centering
    \includegraphics[width=1\textwidth]{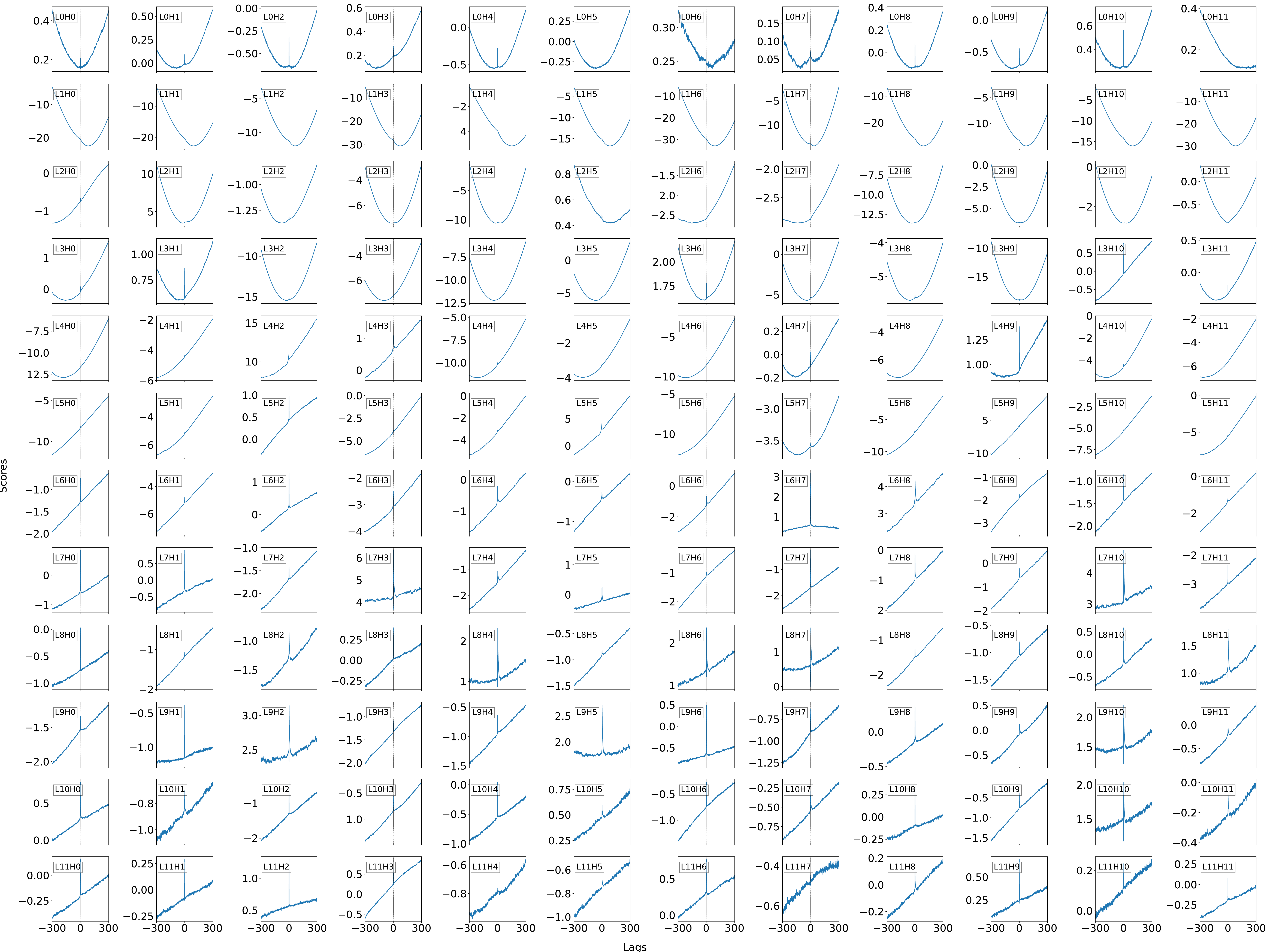}
    \caption{Attention scores as a function of lag for all heads of GPT-2 small after 1000 iterations for Wikitext-103 dataset with baseline amount of positional encoding. \label{fig:crp_all_GPT2-small_1000_1}} 
\end{figure*}

\begin{figure*}[h!]
    \centering
    \includegraphics[width=1\textwidth]{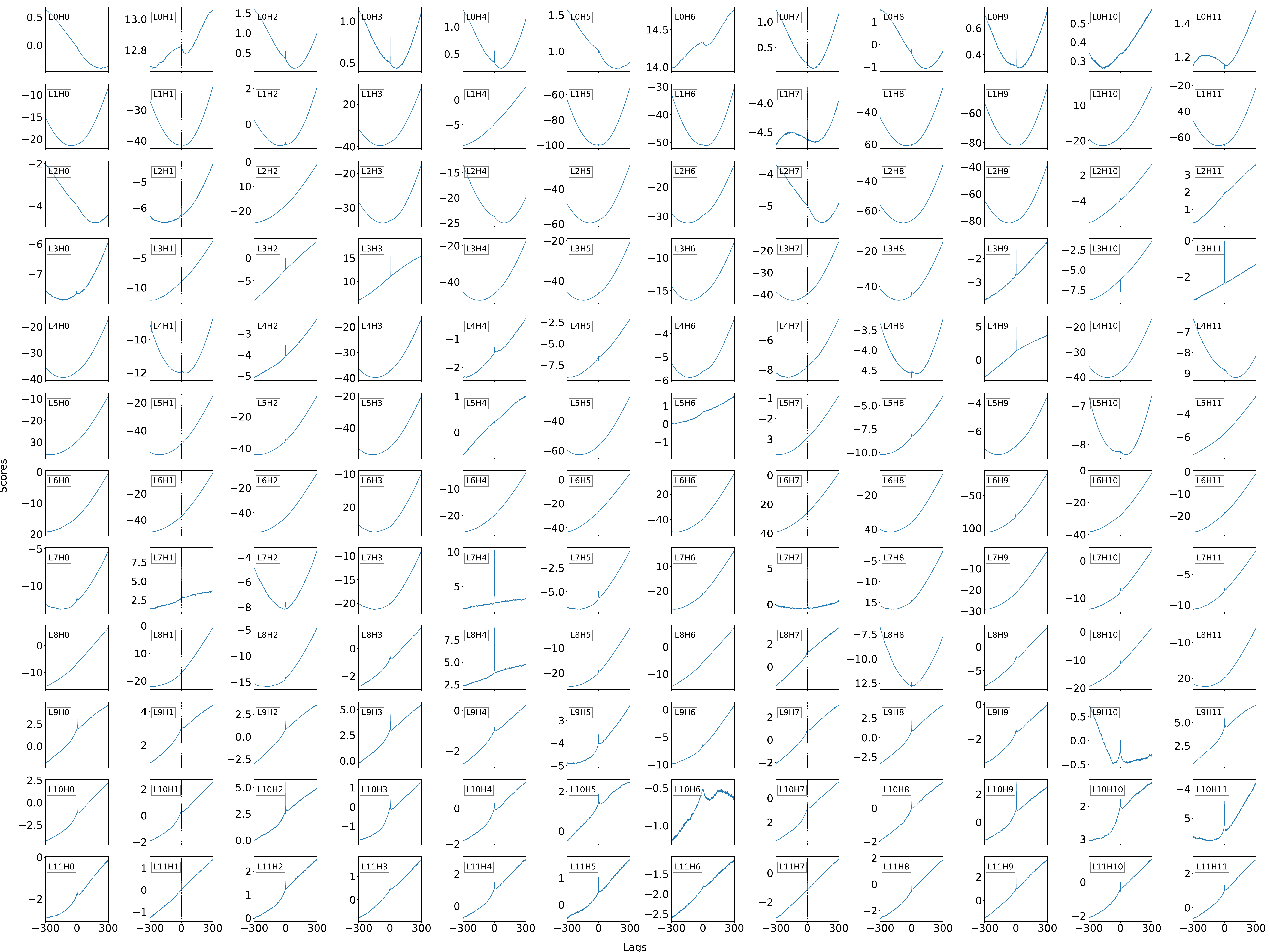}
    \caption{Attention scores as a function of lag for all heads of GPT-2 small after 10000 iterations for FineWeb-1B dataset with baseline amount of positional encoding. \label{fig:crp_all_GPT2-small_1000_1B}} 
\end{figure*}

\begin{figure*}[h!]
    \centering
    \includegraphics[width=1\textwidth]{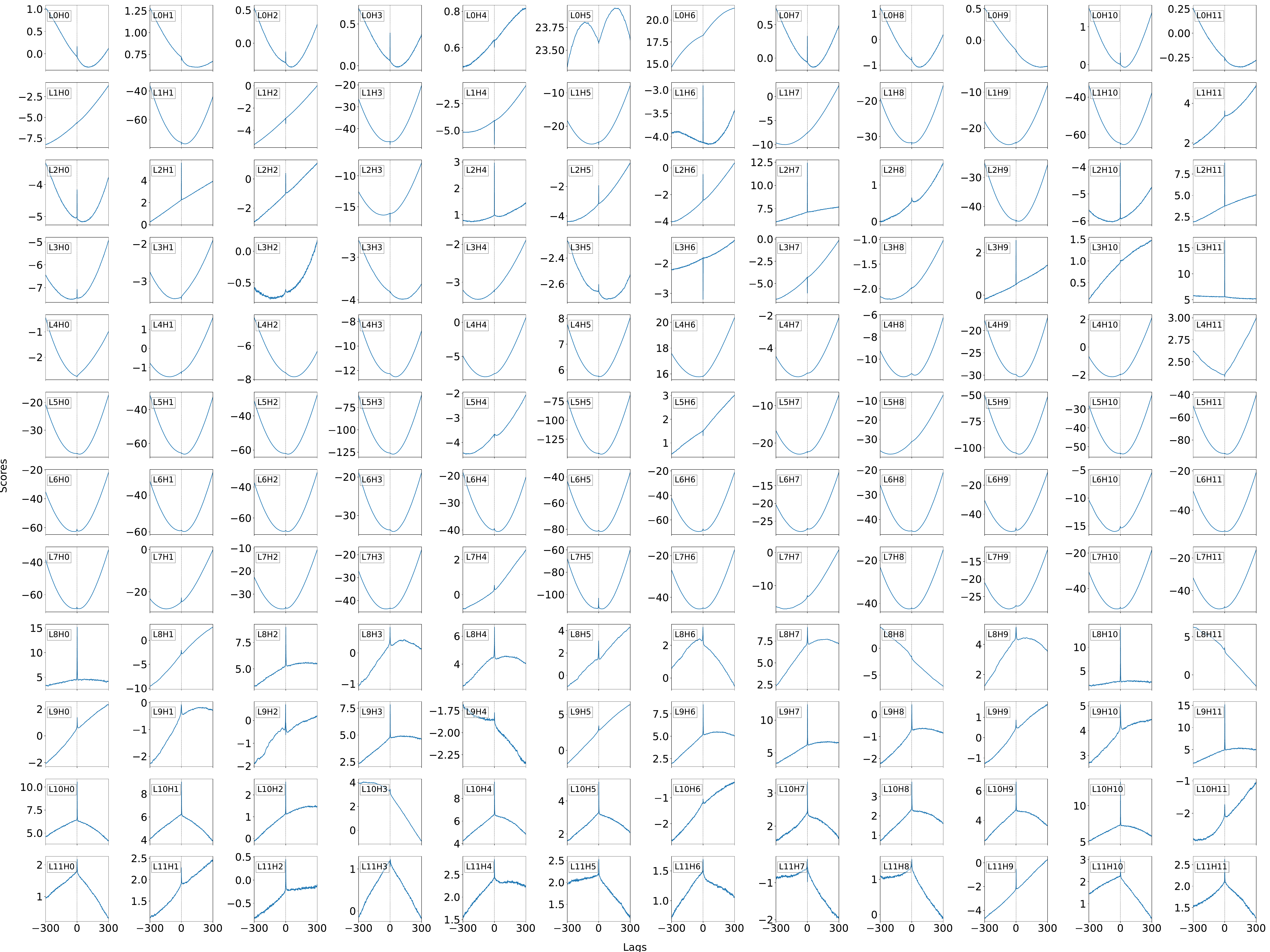}
    \caption{Attention scores as a function of lag for all heads of GPT-2 small after 10000 iterations for FineWeb-10B dataset with baseline amount of positional encoding. \label{fig:crp_all_GPT2-small_1000_10B}} 
\end{figure*}

\end{document}